%% file: 0_main.tex
\documentclass[10pt,twocolumn,letterpaper]{article}

\usepackage[pagenumbers]{iccv} %

\input{0_macros}

\definecolor{iccvblue}{rgb}{0.21,0.49,0.74}
\usepackage[pagebackref,breaklinks,colorlinks,allcolors=iccvblue]{hyperref}

\def\confName{ICCV}
\def\confYear{2025}

\title{RobustSplat: Decoupling Densification and Dynamics for Transient-Free 3DGS}

\author{
    Chuanyu Fu$^{1}$ \quad
    Yuqi Zhang$^{2,3}$ \quad
    Kunbin Yao$^{1}$ \quad
    Guanying Chen$^{1,4}\footnotemark[1]$ \quad
    Yuan Xiong$^{1,4}$ \\
    Chuan Huang$^{3,2}$ \quad 
    Shuguang Cui$^{3,2}$ \quad
    Xiaochun Cao$^{1,4}\footnotemark[1]$ \vspace{0.3em} \\
    {\normalsize $^1$Sun Yat-sen University}\quad{\normalsize $^2$FNii-Shenzhen}\quad{\normalsize $^3$SSE, CUHKSZ}\\{\normalsize $^4$Guangdong Key Laboratory of Information Security Technology}  
}

\begin{document}

\twocolumn[{
\renewcommand\twocolumn[1][]{#1}
\maketitle

\begin{center}
    \captionsetup{type=figure}
    \makebox[0.17\textwidth]{\small Sampled Inputs}
    \makebox[0.20\textwidth]{\small Test View GT}
    \makebox[0.20\textwidth]{\small Ours}
    \makebox[0.20\textwidth]{\small SpotLessSplats}
    \makebox[0.20\textwidth]{\small WildGaussians}
    \\
    \includegraphics[width=\textwidth]{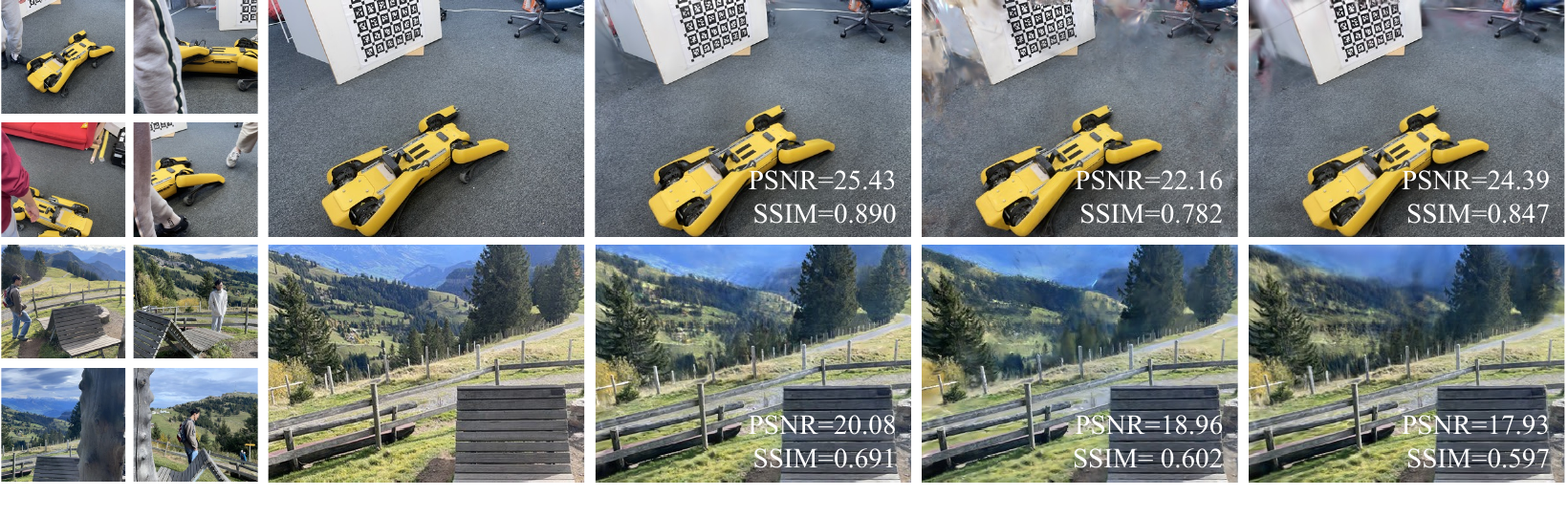}
    \\
    \vspace{-1.05em}
    \captionof{figure}{We propose a robust solution, \emph{RobustSplat}, to handle 3DGS optimization in in-the-wild scenes. Compared with existing approaches, our method significantly reduces artifacts and delivers superior performance, yielding cleaner and more reliable results.
    \label{fig:teaser}
    }
\end{center}
}]

\footnotetext[1]{Corresponding author.}

\input{1_abstract}

\input{2_intro}

\input{3_relatedwork}

\input{4_method}

\input{5_experiments}

\input{6_conclusions}

\clearpage

\input{7_acknowledgements}
{
    \small
    \bibliographystyle{ieeenat_fullname}
    \bibliography{ref}
}

\appendix

\input{8_supp}

\end{document}

%% file: 0_macros.tex
\usepackage{graphicx}
\usepackage{amsmath}
\usepackage{amssymb}

\usepackage{xcolor}
\usepackage{enumitem}
\usepackage{xspace}
\usepackage{booktabs}
\usepackage{colortbl}
\usepackage{multirow}
\usepackage{makecell}
\usepackage{lipsum} %
\usepackage{gensymb} %
\usepackage[page]{appendix} %
\usepackage[accsupp]{axessibility}  %

\newcommand{\Tref}[1]{Table~\ref{#1}}

\newcommand{\fref}[1]{Fig.~\ref{#1}}
\newcommand{\Fref}[1]{Figure~\ref{#1}}

\usepackage[labelsep=period]{caption}
\renewcommand{\paragraph}[1]{\vspace{0.2em}\noindent \textbf{#1 \hspace{0.2em}}}

\definecolor{MyDarkRed}{rgb}{0.56, 0.16, 0.16}
\definecolor{MyBrightRed}{rgb}{0.86, 0.16, 0.16}
\definecolor{MyDarkBlue}{rgb}{0.16, 0.16, 0.66}

\newcommand{\MethodName}{RobustSplat\xspace}
\newcommand{\delayedGS}{delayed Gaussian growth\xspace}
\newcommand{\DelayedGS}{Delayed Gaussian Growth\xspace}
\newcommand{\coarsetofine}{scale-cascaded mask bootstrapping\xspace}

\newcommand{\model}{\mathcal{G}}
\newcommand{\element}{g}
\newcommand{\mean}{\mu}
\newcommand{\covariance}{\boldsymbol{\Sigma}}
\newcommand{\opacity}{\alpha}
\newcommand{\sh}{sh}
\newcommand{\gcolor}{\mathbf{c}}
\newcommand{\loss}{\mathcal{L}}

%% file: 1_abstract.tex
\begin{abstract}

    3D Gaussian Splatting (3DGS) has gained significant attention for its real-time, photo-realistic rendering in novel-view synthesis and 3D modeling. 
    However, existing methods struggle with accurately modeling scenes affected by transient objects, leading to artifacts in the rendered images.
    We identify that the Gaussian densification process, while enhancing scene detail capture, unintentionally contributes to these artifacts by growing additional Gaussians that model transient disturbances.
    To address this, we propose RobustSplat, a robust solution based on two critical designs. 
    First, we introduce a delayed Gaussian growth strategy that prioritizes optimizing static scene structure before allowing Gaussian splitting/cloning, mitigating overfitting to transient objects in early optimization. 
    Second, we design a scale-cascaded mask bootstrapping approach that first leverages lower-resolution feature similarity supervision for reliable initial transient mask estimation, taking advantage of its stronger semantic consistency and robustness to noise, and then progresses to high-resolution supervision to achieve more precise mask prediction.
    Extensive experiments on multiple challenging datasets show that our method outperforms existing methods, clearly demonstrating the robustness and effectiveness of our method. 
    Our project page is \href{https://fcyycf.github.io/RobustSplat/}{https://fcyycf.github.io/RobustSplat/}.
\end{abstract}

%% file: 2_intro.tex
\begin{figure*}[ht] \begin{center}
    \includegraphics[width=\textwidth]{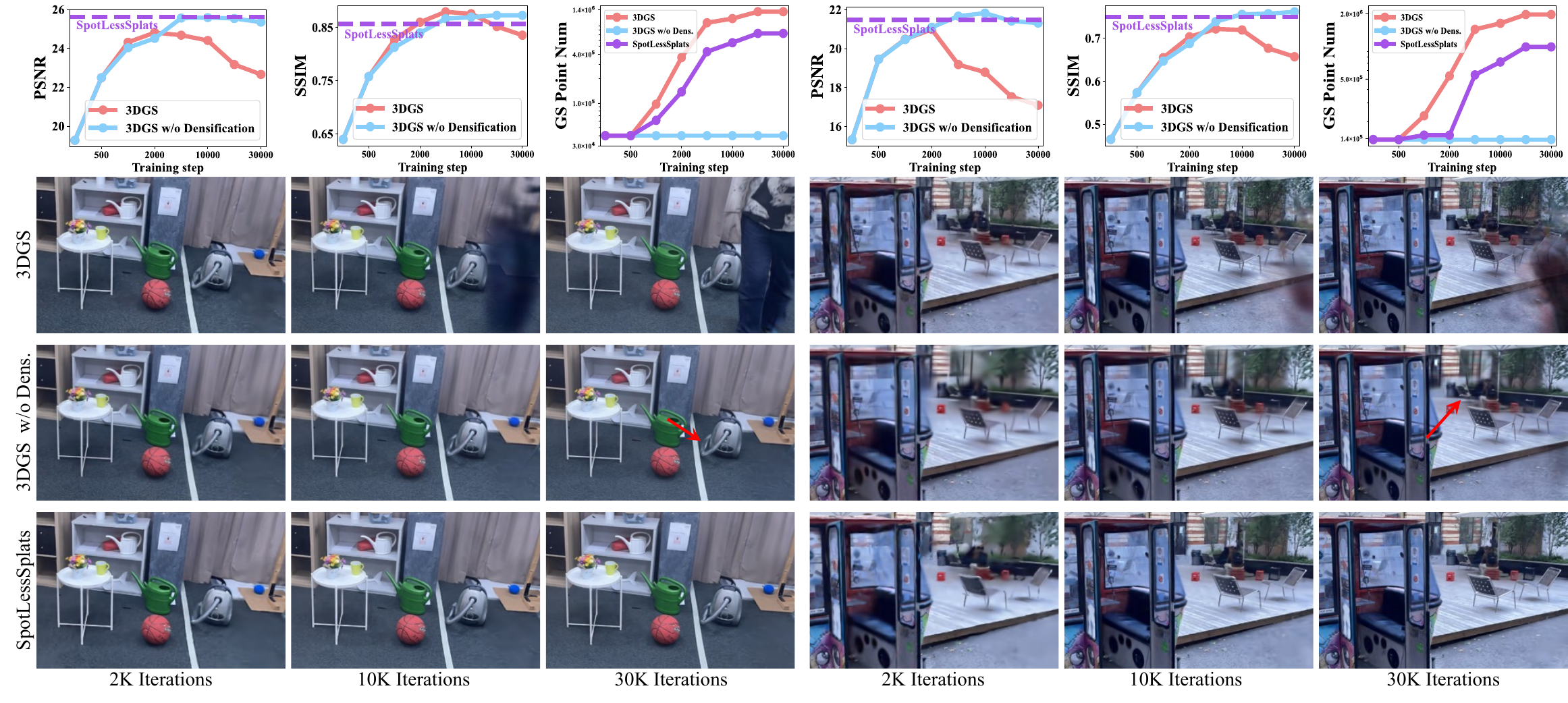}\\
    \vspace{-6pt}
    \makebox[0.48\textwidth]{\small(a) Comparison on Indoor Scene \emph{Corner}}
    \makebox[0.48\textwidth]{\small(b) Comparison on Outdoor Scene \emph{Patio}}
    \caption{\textbf{Analysis of Gaussian densification in transient object fitting.} 
    As training progresses, vanilla 3DGS~\cite{kerbl20233d} suffers from performance degradation and exhibits artifacts due to the increasing number of Gaussians. Disabling Gaussian densification notably improves the results, even achieving performance comparable to the recent robust method SpotLessSplats~\cite{sabour2024spotlesssplats}. 
    Despite producing transient-free rendering, \emph{3DGS w/o densification} struggles to recover fine details in regions with sparse Gaussian initialization (highlighted by \textcolor{MyDarkRed}{red arrows}).}
    \label{fig:dens_analysis}
\end{center} \end{figure*}

\section{Introduction}
\label{sec:intro}

Significant advancements have been made recently in novel-view synthesis and 3D reconstruction from multi-view images \cite{mildenhall2020nerf,tewari2021advances,wu2024recent}. 
Among these, 3D Gaussian Splatting (3DGS) stands out as an effective approach, enabling real-time and realistic rendering~\cite{kerbl20233d}. 
The optimization of 3DGS starts from a sparse set of points obtained through Structure-from-Motion (SfM), and adaptively controls the number and density of Gaussians to create an accurate 3D representation.
To capture fine details, the Gaussians will be split or cloned when the accumulated gradient magnitude of their centered position exceeds a predefined thresholds, 

However, existing methods often assume static scene conditions, an assumption frequently violated in real-world scenarios containing transient objects. 
This mismatch breaks the multi-view consistency requirement, leading to severe artifacts and degraded reconstruction quality~\cite{kulhanek2024wildgaussians}.

\paragraph{Challenges}
The key challenge lies in accurately detecting and filtering motion-affected regions across different images. Existing approaches primarily follow three paradigms:
(1) category-specific semantic masking (e.g., humans and vehicles), which struggles to generalize to diverse transient objects;
(2) uncertainty-based masking by considering uncertainty in minimizing photometric reconstruction loss, but often fails to reliably predict motion masks~\cite{martin2021nerf}; and
(3) learning-based motion masking, where an MLP predicts motion masks using image features (e.g., DINO features~\cite{oquab2023dinov2}) as input and is supervised by photometric residuals~\cite{sabour2024spotlesssplats} or feature similarity~\cite{kulhanek2024wildgaussians,goli2024romo} between captured and rendered images.

While learning-based methods have shown strong performance in transient-free 3DGS optimization, they face critical limitations. 
In the early stages of training, the 3DGS representation is under-optimized, resulting in over-smooth renderings with large photometric residuals and weak feature similarity in both dynamic and static regions. 
Using these unreliable signals as supervision for mask estimation leads to inaccurate transient masks, with small masks failing to remove transients and causing artifacts, while overly smooth early reconstructions misclassify static regions, hindering optimization and resulting in under-reconstruction, as shown in~\fref{fig:teaser}.

\paragraph{Analysis} To mitigate these issue, two critical aspects need to be considered. 
First, the optimization of 3DGS should be explicitly constrained from overfitting to transient regions without accurate transient mask the during initial optimization phases.
Second, the mask supervision in early iterations should be designed to be more tolerant to under-reconstructed regions in the early optimization to allow reconstruction of static regions.

Through a detailed analysis of the 3DGS method, we identify that the Gaussian densification process (which, by default, begins after 500 iterations) enhances scene detail capture but unintentionally introduces artifacts (see~\fref{fig:dens_analysis}). 
Initially, 3DGS fits the static parts of the scene well; however, as densification progresses, it tends to overfit dynamic regions, resulting in artifacts in areas influenced by moving objects. 
Surprisingly, we find that \emph{explicitly disabling the densification process in vanilla 3DGS effectively mitigates these artifacts}, yielding results comparable to SpotLessSplats~\cite{sabour2024spotlesssplats} without requiring any specialized design. 

This is because, without densification, the image reconstruction loss provides limited positional gradients for 3D Gaussians, primarily optimizing their shape and color instead. As a result, the initially placed Gaussians remain stable in position, reducing the risk of overfitting to transient elements.  
However, the absence of densification leads to an insufficient number of Gaussians to represent fine details, causing the rendered images to appear overly smooth in regions with sparse point initializations.

\paragraph{The Proposed Solution}
Building on our analysis, we propose a simple yet effective method, called \emph{RobustSplat}, for optimizing 3DGS in in-the-wild scenes. 
Our method introduces two key designs. 
First, we propose a \emph{\delayedGS} strategy that prioritizes reconstructing the global structure of the 3D scene while explicitly avoiding premature fitting to dynamic regions.
Second, to improve the mask supervision signal for under-reconstructed regions while preserving sensitivity to transient areas, we introduce a \emph{\coarsetofine} approach. This approach progressively increase the supervision resolution, leveraging the observations that low-resolution features capture global consistency more effectively and suppressing local noise in early optimization stages.

In summary, our key contributions are:  
\begin{itemize}[itemsep=0pt,parsep=0pt,topsep=2bp]
    \item We analyze how the 3DGS densification process contributes to artifacts caused by transient objects, offering new insights for improving the optimization of distractor-free 3DGS.
    \item We propose \emph{RobustSplat}, a robust approach that integrates the \delayedGS strategy and \coarsetofine to effectively reduce the impact of dynamic objects during 3DGS optimization.
    \item We demonstrate that our approach outperforms state-of-the-art methods with a simple yet effective design.
\end{itemize}

%% file: 3_relatedwork.tex
\section{Related Work}
\label{sec:related_works}

\paragraph{Novel View Synthesis} 
Neural radiance field (NeRF)~\cite{mildenhall2020nerf}, as a representative approach for novel view synthesis, is widely recognized for its highly realistic rendering capabilities~\cite{tewari2021advances, tewari2020state, tang2022compressible, xie2022neural}. 
Many follow-up NeRF-based methods have introduced numerous enhancements in terms of efficiency~\cite{muller2022instant, chen2022tensorf, fridovich2022plenoxels} and performance~\cite{barron2022mip, lu2023large, mari2022sat, oechsle2021unisurf, wang2021neus, wu2022scalable, xu2023grid, yu2022monosdf, zhang2024aerial,yang2023freenerf}, achieving highly effective results. Recently, a novel explicit representation, 3D Gaussian Splatting (3DGS)~\cite{kerbl20233d}, has sparked considerable attention for its transformative impact on novel view synthesis methods due to its real-time rendering capability~\cite{yu2024mip, guedon2024sugar, lin2024vastgaussian, zhou2024hugs}.

\paragraph{Robustness in NeRF}
The vanilla NeRF assumes a static scene, but this assumption often fails with in-the-wild situations, where unconstrained images inevitably include lighting variations and dynamic/transient objects. 
NeRF-W~\cite{martin2021nerf} introduces an appearance embedding for exposure and transient modeling, which has been widely used~\cite{yang2023cross,chen2022hallucinated}. 
For distractors removal, it uses MLPs to predict uncertainty and following methods~\cite{lee2023semantic, ren2024nerf} introduce features from large pre-trained models~\cite{caron2021emerging, oquab2023dinov2} to improve robustness.
Another branch, represented by RobustNeRF~\cite{sabour2023robustnerf}, utilizes image residuals to predict binary masks for dynamic objects, filtering them out during training~\cite{otonari2024entity, chen2024nerf}.
Moreover, $D^{2}$-NeRF~\cite{wu2022d} decouples a dynamic scene into three fields, including static field, dynamic field, and non-static shadow field.

\paragraph{Robustness in 3DGS}
Unlike NeRF, which uses a continuous MLP-based implicit representation, 3DGS employs a discrete explicit representation. As a result, many studies~\cite{zhang2024gaussian, wang2024we, darmon2024robust, kulhanek2024wildgaussians,tang2024nexussplats} explore strategies that combine global information of reference images with local Gaussian features for illumination modeling. 
For distractors removal, transient objects are typically filtered out using masks~\cite{xu2024wild, ungermann2024robust,wang2024distractor,xu2024splatfacto,dahmani2024swag,wang2024uw,bao2024distractor}.

To handle transient objects, WildGaussians~\cite{kulhanek2024wildgaussians} incorporate the DINO~\cite{oquab2023dinov2} features to predict uncertainty, which is then converted into a mask.
Robust3DGaussians~\cite{ungermann2024robust} enhances the predicted mask by leveraging SAM~\cite{kirillov2023segment}.
SpotLessSplats~\cite{sabour2024spotlesssplats} leverages features from Stable Diffusion~\cite{rombach2022high}, designing two clustering strategies for mask prediction. 
T-3DGS~\cite{pryadilshchikov2024t} introduces an unsupervised transient detector based on a consistency loss and a video object segmentation module to track objects in the videos.

More recently, DeSplat~\cite{wang2024desplat} decompose the 3DGS scenes into a static 3DGS and per-view transient 3DGS by only minimizing the photometric loss. HybridGS~\cite{lin2024hybridgs} instead combines 3DGS with per-view 2D image Gaussians to decouple dynamics and statics.
DAS3R~\cite{xu2024das3r} and RoMo~\cite{goli2024romo} proposed to estimate motion mask for dynamic videos by making use of the temporal consistency constraints, which cannot be directly applied to a set of unordered images.
Different from existing methods, we analyse the densification process of 3DGS and propose a simple yet effective solution based on the \delayedGS and \coarsetofine to reliably remove the effects of trainsient objects.

\paragraph{Optimization in Densification and Regularization} 
There are prior works aiming to improve the densification and optimization process of 3DGS~\cite{zhang2024fregs,fang2024mini,bulo2024revising,hyung2024effective,bulo2024revising}.
For example, several methods~\cite{ye2024absgs, zhang2024pixel,yu2024gaussian} have analyzed the gradient computation process and identified issues such as gradient collision or averaging, which lead to suboptimal reconstruction quality. 
RAIN-GS~\cite{jung2024relaxing} investigates alternative initialization strategy for 3DGS without relying on COLMAP SfM.
These methods does not consider the effect of transient objects.
In this work, we analyse and leverage of the behaviors of Gaussian densification in context of transient-free 3D reconstruction.

%% file: 4_method.tex
\section{Method}
\label{sec:method}

\subsection{Overview}
\label{sub:Overview}
Given casually captured multi-view posed images with transient objects, our goal is to optimize a clean 3D Gaussian splatting representation that enable distractor-free novel-view synthesis.
Our approach builds upon recent robust 3DGS methods that jointly optimize 3D representation and transient object masks during training~\cite{sabour2024spotlesssplats}.
The transient masks selectively filter dynamic regions in images, while improve scene modeling by providing more accurate supervision for mask MLP optimization. 

However, this interdependence can lead to instability in early training.
On one hand, if the masks are too small, they fail to filter all transient regions, causing newly generated Gaussians to fit transient objects. This makes it difficult to remove artifacts in later stages. 
On the other hand, the static scene reconstruction is often overly smooth in the early stage, which will misguide the mask MLP into incorrectly classifying static regions as dynamic, hindering their reconstruction and leading to under-representation of static content.

\begin{figure}[tb] \centering
    \includegraphics[width=0.48\textwidth]{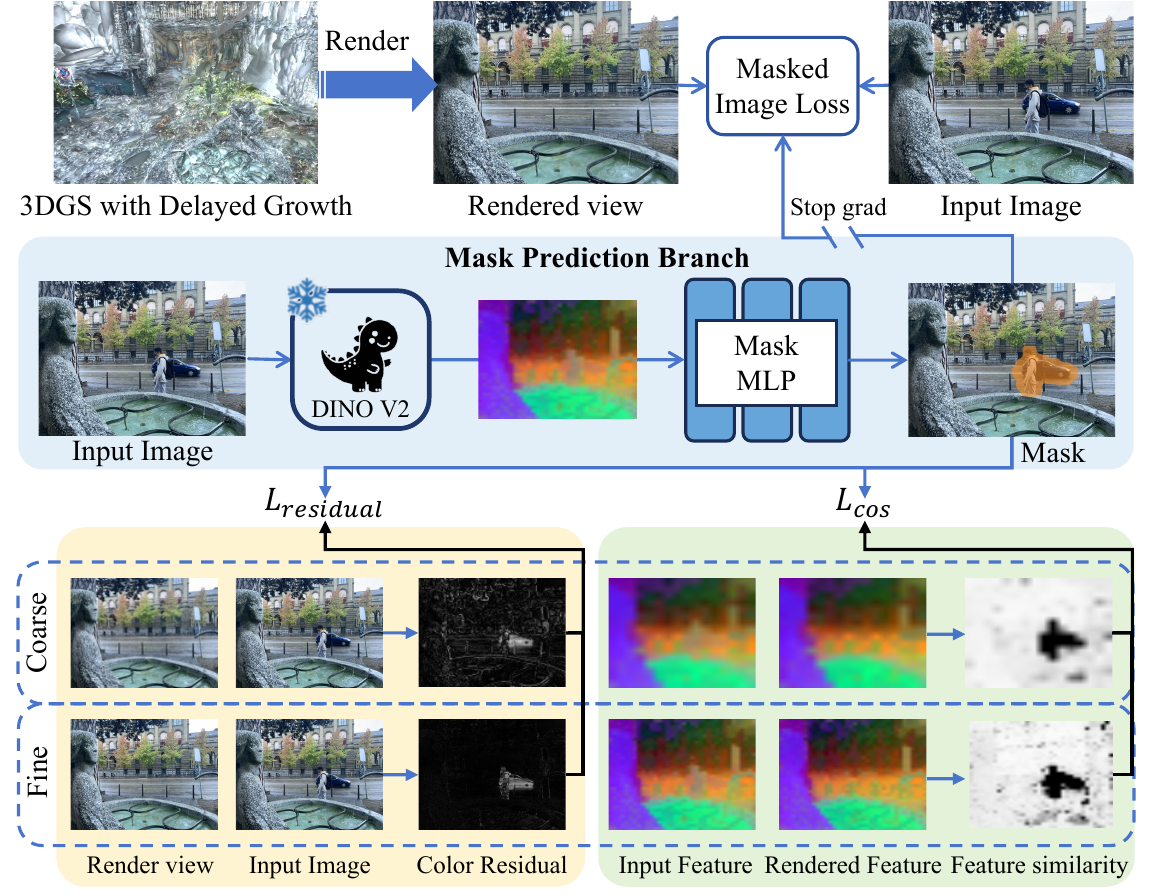}
    \caption{Overview of the proposed \MethodName.
    } 
    \label{fig:pipeline}
\end{figure}

To address these challenges, we introduce two effective designs (see~\fref{fig:pipeline}).
First, we introduce a \delayedGS strategy to postpones the Gaussian densification process to prevent fitting transient objects in the early stage. 
Second, we propose a \coarsetofine approach to refine mask predictions over time, reducing the misclassification of static regions as transient and improving the optimization of static content.

\subsection{3DGS with Transient Mask Estimation}
\paragraph{3D Gaussian Splatting} 
We represent the scene as a set of 3D Gaussians $\model{=} \{ \element_i \}_{i=1}^N$, where each Gaussian primitive $\element_i$ has learnable parameters including mean position $\mean_i$, covariance matrix $\covariance_i$ for shape, opacity $\opacity_i$, and spherical harmonics coefficients $\sh_i$ for view dependent color~\cite{kerbl20233d}. 
For novel view synthesis, 3D Gaussians are projected to 2D and rendered by differentiable rasterization using alpha blending~\cite{zwicker2001surface}. The final pixel color $\gcolor_k$ is computed via alpha blending:
\begin{equation}
\gcolor_k = \sum^N_{i=1} \gcolor_i\,\alpha_i\,\model^{2D}_i \prod_{j=1}^{i-1} (1 - \alpha_j\,\model^{2D}_j),
\end{equation}
where $\gcolor_i$ is the color computed from spherical harmonics coefficients with the view direction.

The 3DGS is optimized by minimizing the L1 loss and SSIM loss between the rendered and the captured images:
\begin{equation}
\loss = ( 1 - \lambda ) \loss_1 + \lambda \loss_\textrm{D-SSIM}.
\end{equation}
During optimization, adaptive density control periodically clones/prunes Gaussians based on accumulated positional gradient magnitudes.

\begin{figure}[t] \centering
    \makebox[0.116\textwidth]{\footnotesize }
    \makebox[0.116\textwidth]{\footnotesize DINOv2}
    \makebox[0.116\textwidth]{\footnotesize SAM2}
    \makebox[0.110\textwidth]{\footnotesize StableDiffusion}
    \includegraphics[width=0.48\textwidth]{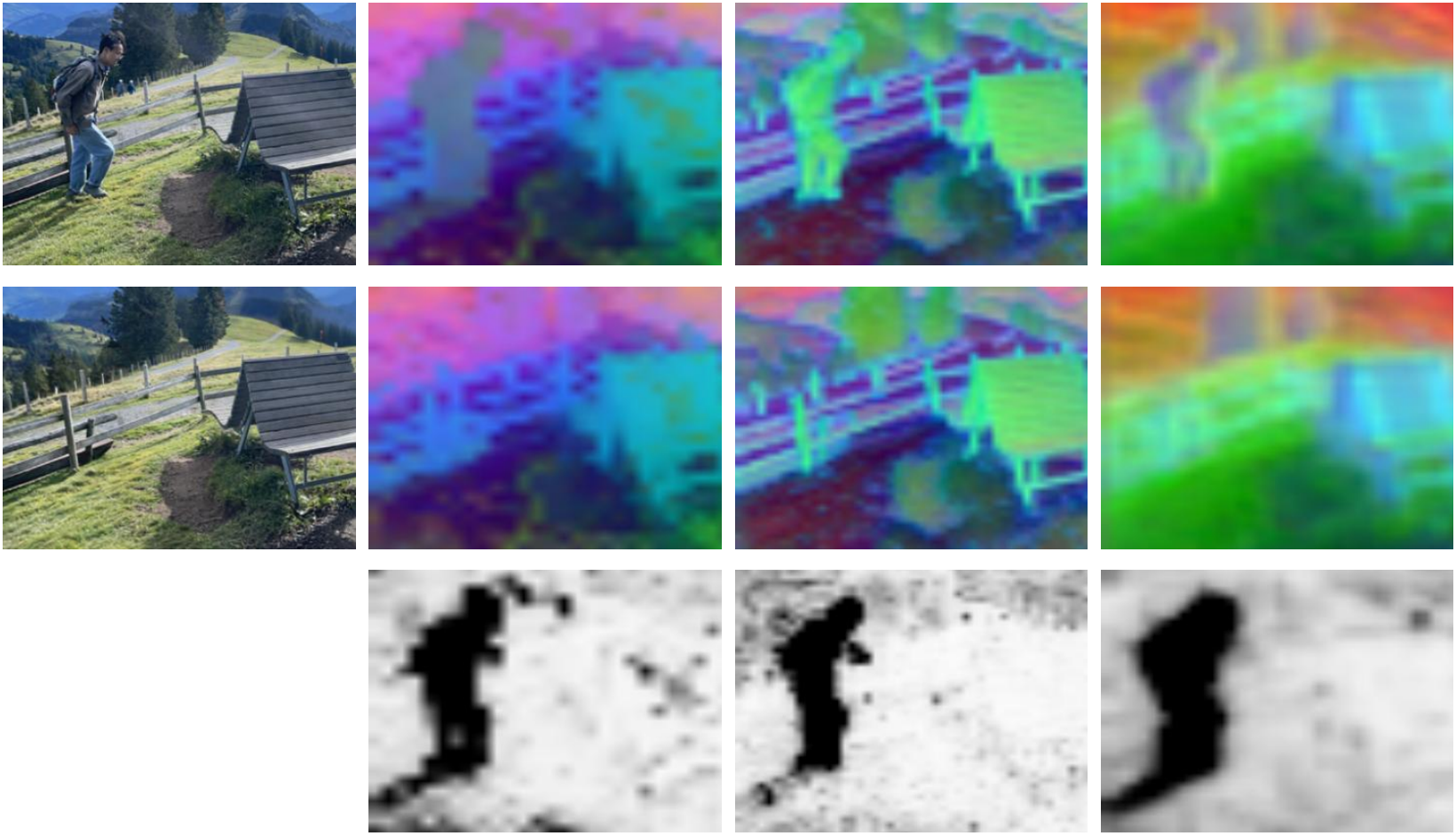}
    \caption{Visualization of DINOv2, SAM2, and SD features via PCA. The last row compares the cosine similarity maps between features of the ground-truth and rendered image.} \label{fig:feat_compare}
\end{figure}

\paragraph{Transient Mask Estimation}
To deal with transient objects, following recent work~\cite{sabour2024spotlesssplats,kulhanek2024wildgaussians}, we predict per-image transient masks $M_t$ using an MLP conditioned on image features $f_t$:
\begin{equation}
M_t = \textrm{Sigmoid}(\textrm{MLP}_\textrm{mask}(f_t)).
\end{equation}
The estimated mask is then used to apply a masked photometric loss that excludes transient regions.

Recent works utilize features containing strong semantic information as MLP inputs (e.g., DINOv2~\cite{oquab2023dinov2,kulhanek2024wildgaussians}, StableDiffusion~\cite{rombach2022high,sabour2024spotlesssplats}, SAM~\cite{kirillov2023segment,goli2024romo}). 
Our preliminary experiments found that StableDiffusion feature provided stronger semantic information, but it is computationally expensive to extract the feature. Despite SAM features are better at produce mask with more accurate boundary, it struggle to locate the shadow regions, which produce incomplete mask prediction, as shown in~\fref{fig:feat_compare}.
We employ the DINOv2 features as input to the MLP as it maintains a good balance of computational efficiency and semantic extraction ability.

\paragraph{Optimization of Mask MLP}
The optimization of the MLP weight requires appropriate supervision. 
We adopt the image robust loss $\mathcal{L}_\textrm{residual}$ based the image residual information introduced in \cite{sabour2024spotlesssplats} as one of the supervision.

To better leverage deep high-dimensional feature information extracted from images, which have different properties as the image residual, we adopt a feature robust loss $\mathcal{L}_\textrm{cos}$ utilizing the information of feature similarity between the rendered and captured images.
Specifically, we extract DINOv2 features of the real image $f_t$ and rendered image $f_t'$, and compute their cosine similarity map. 
Then we convert the cosine similarity map to be in the value range of $[0, 1]$ following \cite{kulhanek2024wildgaussians}:
\begin{align}
    \label{eq:}
    M_\textrm{cos} = \textrm{max} \left( 2cos\left ( f_t,f_t' \right ) -1,0 \right),
\end{align}
where $M_\textrm{cos}$ will be $1$ is the feature cosine similarity is 1, and it will be $0$ if the similarity is less than $0.5$.
Then the feature robust loss is expressed as:
\begin{equation}
    \mathcal{L}_\textrm{cos} = \left \| M_t -  M_\textrm{cos} \right \|.
\end{equation}
The MLP is optimized using the following loss:
\begin{equation}
    \mathcal{L}_\textrm{MLP} = \lambda_\textrm{residual} \mathcal{L}_\textrm{residual} + \lambda_\textrm{cos} \mathcal{L}_\textrm{cos},
\end{equation}
where $\lambda_\textrm{residual}$, $\lambda_\textrm{cos}$ are the corresponding weights for image robust supervision and feature robust loss, respectively.

\subsection{Delayed Gaussian Growth for Mask Learning}

Motivated by our observation that disabling Gaussian densification in 3DGS significantly improves the learning of low-frequency static components, we introduce a delayed Gaussian growth strategy, modifying 3DGS~\cite{kerbl20233d} to defer Gaussian densification during optimization.

\begin{figure}[tb] \centering
    \includegraphics[width=0.235\textwidth]{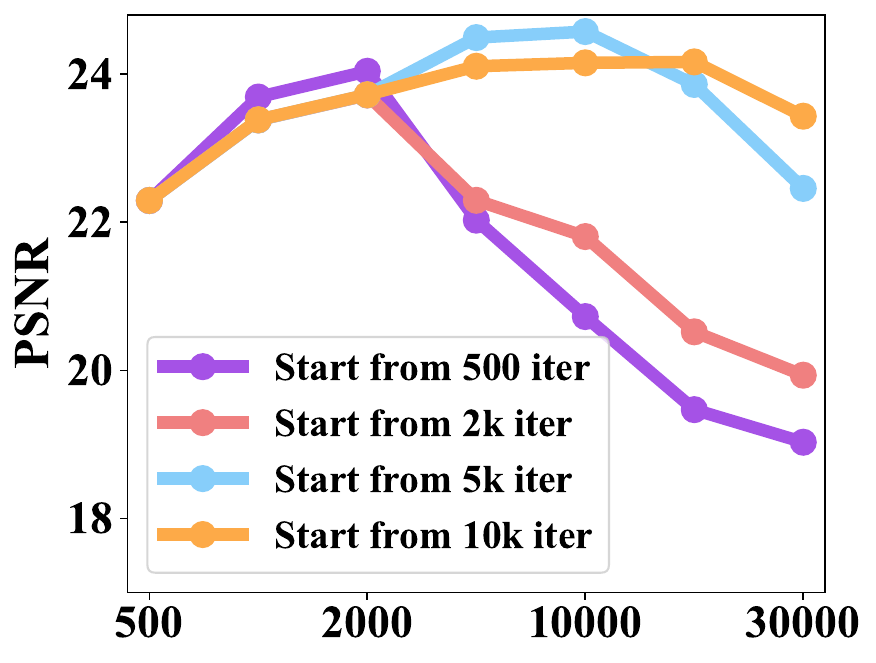}
    \includegraphics[width=0.235\textwidth]{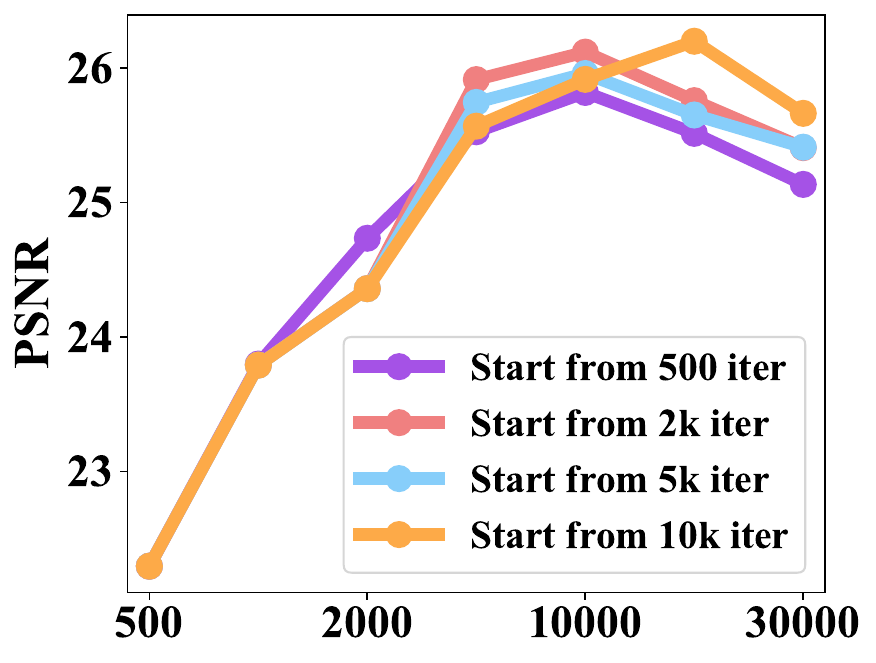}
    \\
    \vspace{-6pt}
    \makebox[0.235\textwidth]{\small (a) w/o robust mask} 
    \makebox[0.235\textwidth]{\small (b) with robust mask} 
    \caption{Effects of start iteration of Gaussian densification with and without the transient mask learning. } \label{fig:analysis_start}
\end{figure}

\paragraph{Analysis of Delayed Gaussian Growth} 
To evaluate the impact of the Gaussian densification start time in 3DGS, we vary the initial densification iteration while keeping the densification interval fixed at 10K iterations. 
As shown in~\fref{fig:analysis_start}~(a), delaying densification allows 3DGS to focus on reconstructing the static scene during the early training stages. 
However, once densification begins, newly introduced Gaussians tend to fit transient objects, leading to a decline in PSNR metrics. Notably, models with earlier densification exhibit worse performance, indicating that premature densification promotes transient object fitting. 
These results suggest that postponing densification helps the model better capture the static components before accommodating dynamic elements.

\paragraph{Mask Learning with Delayed Gaussian Growth} 
To mitigate transient artifacts caused by uncontrolled Gaussian growth, we incorporate transient mask learning into the delayed densification process. 
As shown in~\fref{fig:analysis_start}~(b), this integration significantly improves reconstruction accuracy by leveraging mask predictions to regulate Gaussian expansion.
By leveraging mask predictions to regulate Gaussian expansion, this approach effectively suppresses transient artifacts and enhances scene fidelity.
In particular, variants with a later densification start achieve more accurate results. These results demonstrate that transient mask learning and delayed densification work collaboratively to enhance the stability and accuracy of 3DGS optimization.

\paragraph{Mask Regularization at Early Stage} 
The timing of applying transient mask filtering in 3DGS is a critical aspect. In the initial training phase, rendered images exhibit low quality with large image residuals and poor feature similarity, leading to inaccurate mask estimation. 
To mitigate this, prior methods either delay mask learning until after a warm-up period (e.g., 1500 iterations)~\cite{kulhanek2024wildgaussians} or employ random mask sampling strategies~\cite{sabour2024spotlesssplats}. 
However, delaying mask application risks incorporating transient objects into the scene, making them harder to remove later.

Thanks to the delayed strategy for Gaussian growth, our approach ensures that early-stage optimization focuses solely on static scenes.
To facilitate the optimization of static regions across the entire scene, we encourage the mask MLP to initially classify all regions as static and gradually filter out transient objects.
To achieve this, we introduce a regularization term into the mask MLP’s supervision:
\begin{equation}
    \loss_\textrm{reg} = e^{(-\frac{i}{\beta_{reg}})}\left \| 1 - M_t \right \|,
\end{equation}
where $i$ is the iteration number of training, and the right term is 1 if $i=0$, and will decrease when $i$ increases. 

The overall loss for mask optimization is expressed as:
\begin{equation}
    \mathcal{L}_\textrm{MLP} = \lambda_\textrm{residual} \mathcal{L}_\textrm{residual} + \lambda_\textrm{cos} \mathcal{L}_\textrm{cos} + \lambda_\textrm{reg} \mathcal{L}_\textrm{reg},
\end{equation}
where $\lambda_\textrm{reg}$ is the corresponding weights for regularization.

\begin{figure}[t] \centering
    \makebox[0.006\textwidth]{\scriptsize}
    \makebox[0.088\textwidth]{\scriptsize Masked GT}
    \makebox[0.088\textwidth]{\scriptsize Rendering}
    \makebox[0.088\textwidth]{\scriptsize GT Feature}
    \makebox[0.088\textwidth]{\scriptsize Rendered Feature}
    \makebox[0.08\textwidth]{\scriptsize Cosine}
    \includegraphics[width=0.48\textwidth]{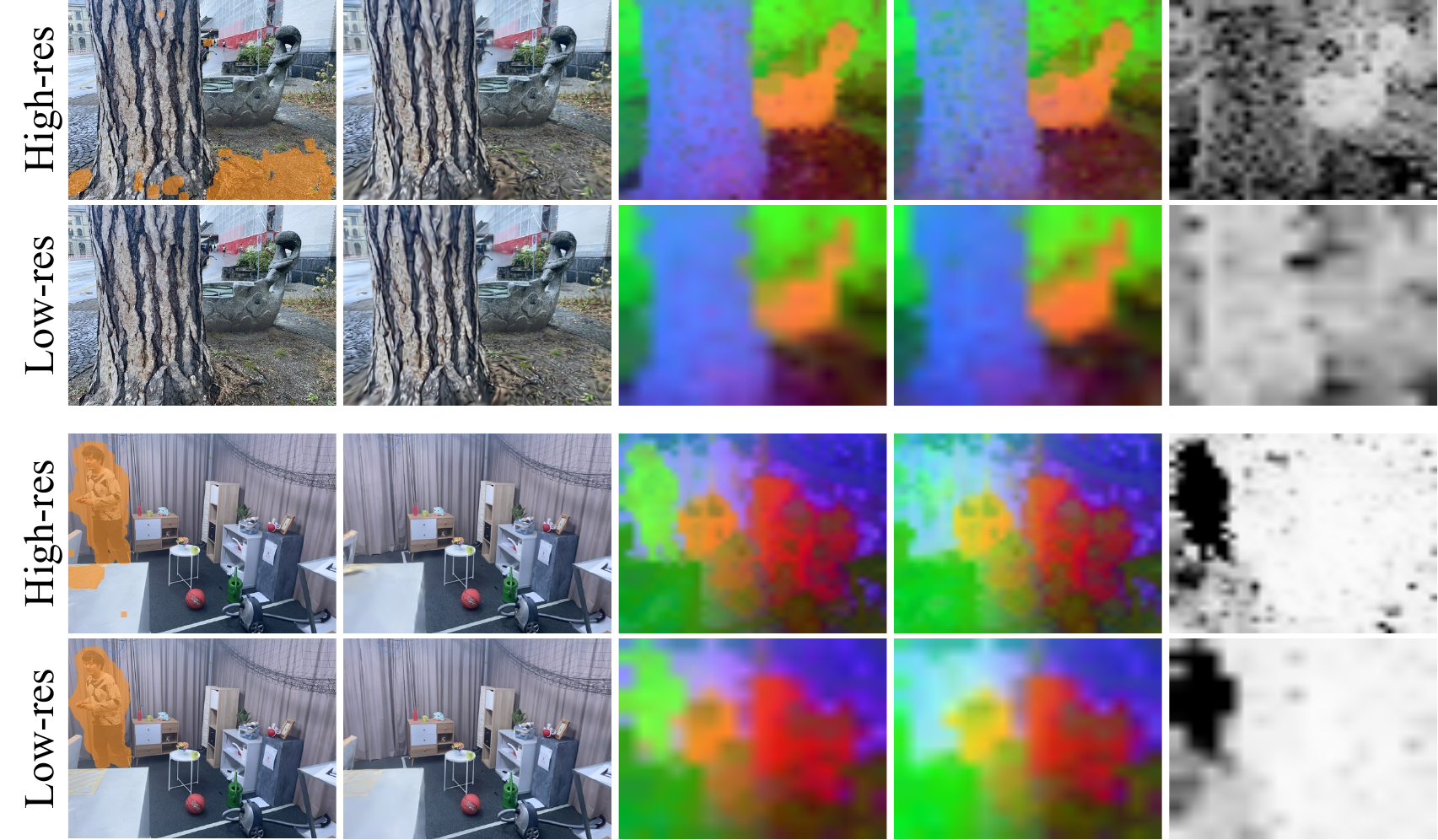}
    \caption{Effects of mask supervisions derived from different resolutions on two scenes. The first column shows input images overlapped with yellow masks predicted by the mask MLP after training with supervisions derived from the corresponding resolutions.} \label{fig:analysis_mask_feature}
\end{figure}

\subsection{Scaled-cascaded Mask Bootstrapping}

While our \delayedGS strategy effectively mitigates the influence of transient regions by focusing optimization on static areas, the under-reconstruction of static scenes remains an issue in the early stages. This problem arises due to the sparsity of the initial Gaussian points, particularly in large-scale unbounded outdoor scenes. 
Consequently, the rendered outputs in these regions appear overly smooth, leading to large image residuals and low feature similarity. This, in turn, causes the mask MLP to misclassify under-reconstructed static areas as dynamic.

\paragraph{Robust Feature Similarity Computation}
To address this, we aim to make the supervision signal more tolerant to under-reconstructed regions in the early optimization phase. We observe that while high-resolution features extracted from high-resolution images provide fine-grained spatial details, they suffer from limited receptive fields and increased sensitivity to local noise. In contrast, low-resolution features capture global consistency more effectively, as each patch integrates broader contextual information, inherently suppressing local noise in feature representations.

As shown in~\fref{fig:analysis_mask_feature}, compared to high-resolution image, low-resolution images naturally suppress fine details, leading to smoother color residuals and features similarity. This suggests that evaluating residuals and feature similarity at a lower resolution during the early stages improves robustness—allowing under-reconstructed static regions to be retained while maintaining sensitivity to transient areas.

\paragraph{Coarse-to-fine Mask Supervision}
Building on this insight, we propose a resolution-cascaded approach that progressively refines mask supervision by transitioning from low-resolution to high-resolution signals. This method helps the mask MLP retain more static regions in the early optimization phase.

Specifically, before the start of Gaussian densification,  we render images with low-resolution from 3DGS to compute low-resolution image residuals and feature consistency to supervise the mask MLP. 
Once densification begins, we switch to high-resolution residuals and cosine similarity between high-resolution features, ensuring finer-grained discrimination of transient and static regions.

\begin{table*}[t] \centering
    \begin{center}
    \captionof{table}{Quantitative comparison on NeRF On-the-go Dataset. The best results are highlighted in \textbf{bold}, and the second in \underline{underline}.}
    \label{table:onthego}
    \input{tables/onthego}
    \end{center}
    \vspace{-6pt}
    \input{images/qual_onthego}
    
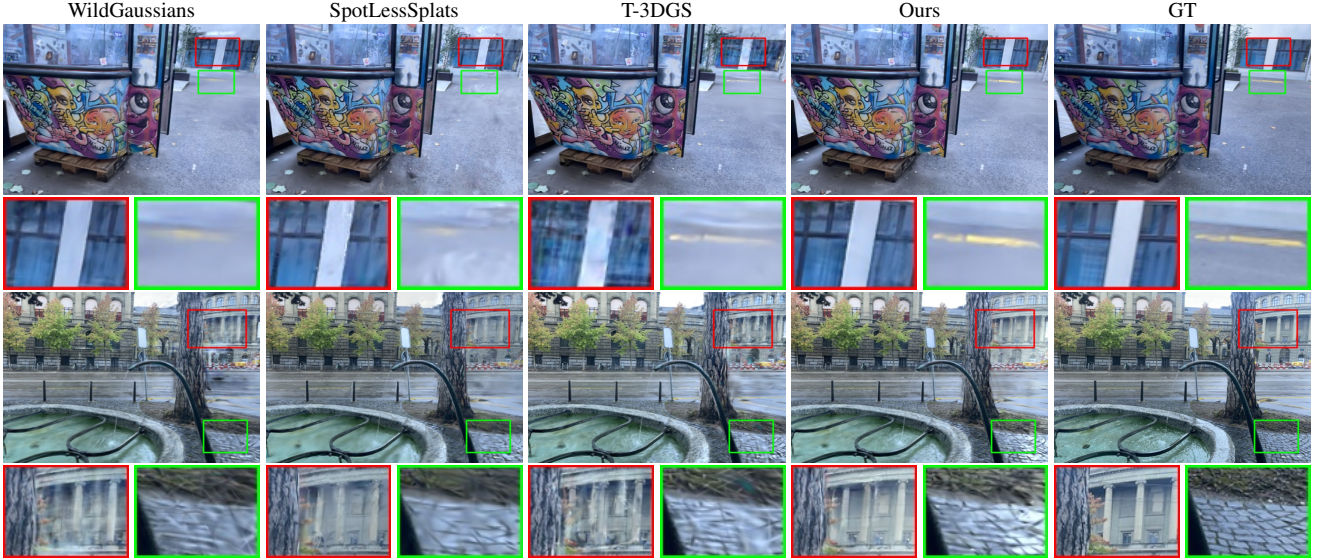
\captionof{figure}{Qualitative results on \emph{Patio-high} and \emph{Fountain} from NeRF On-the-go dataset.}
    \label{fig:qual_onthego}
    \vspace{-1.2em}
\end{table*}

%% file: tables/onthego.tex
\resizebox{\textwidth}{!}{
    \begin{tabular}{l|ccc|ccc|ccc|ccc|ccc|ccc|ccc}
    \toprule
    \multirow{3}{*}{Method} 
    & \multicolumn{6}{c|}{Low Occlusion}  
    & \multicolumn{6}{c|}{Medium Occlusion} 
    & \multicolumn{6}{c|}{High Occlusion} 
    & \multicolumn{3}{c}{\multirow{2}{*}{Mean}}  
    \\
    \multicolumn{1}{l|}{}  
    & \multicolumn{3}{c}{Mountain}  
    & \multicolumn{3}{c|}{Fountain} 
    & \multicolumn{3}{c}{Corner}  
    & \multicolumn{3}{c|}{Patio} 
    & \multicolumn{3}{c}{Spot} 
    & \multicolumn{3}{c|}{Patio-High}  
    \\
     & PSNR & SSIM & LPIPS 
     & PSNR & SSIM & LPIPS 
     & PSNR & SSIM & LPIPS 
     & PSNR & SSIM & LPIPS 
     & PSNR & SSIM & LPIPS 
     & PSNR & SSIM & LPIPS 
     & PSNR & SSIM & LPIPS 
     \\
    \midrule
    3DGS~\cite{kerbl20233d}
    & 19.21 & 0.691 & 0.229 
    & 20.08 & \underline{0.686} & \underline{0.208} 
    & 22.65 & 0.835 & 0.162 
    & 17.04 & 0.713 & 0.232 
    & 18.54 & 0.717 & 0.334 
    & 17.04 & 0.657 & 0.314 
    & 19.09 & 0.717 & 0.248\\
    
    SpotLessSplats~\cite{sabour2024spotlesssplats}
    & 20.67 & 0.670 & 0.282 
    & 20.63 & 0.645 & 0.265 
    & 25.47 & 0.858 & 0.155 
    & \underline{21.43} & 0.803 & 0.171 
    & 23.64 & 0.819 & 0.207 
    & 21.17 & 0.749 & 0.237 
    & 22.17 & 0.757 & 0.220\\
    
    WildGaussians~\cite{kulhanek2024wildgaussians}
    & \underline{20.77} & 0.697 & 0.268 
    & 20.48 & 0.666 & 0.250 
    & 25.21 & 0.865 & 0.136 
    & 21.17 & 0.804 & 0.168 
    & 24.60 & 0.871 & 0.135 
    & 22.44 & 0.802 & 0.184
    & 22.45 & 0.784 & 0.190\\

    Robust3DGaussians~\cite{ungermann2024robust}
    & 19.47 & 0.672 & 0.251 
    & 19.74 & 0.653 & 0.254 
    & 24.41 & 0.869 & 0.118 
    & 16.63 & 0.729 & 0.209 
    & 22.64 & 0.874 & 0.132 
    & 21.56 & 0.799 & 0.174 
    & 22.45 & 0.766 & 0.190
\\
    
    T-3DGS~\cite{pryadilshchikov2024t}
    & 20.62 & \underline{0.703} & \underline{0.223} 
    & \underline{20.83} & 0.681 & 0.218 
    & \underline{26.14} & \underline{0.890} & \underline{0.114} 
    & 20.96 & \underline{0.819} & \underline{0.154} 
    & \underline{25.84} & \underline{0.893} & \underline{0.127} 
    & \underline{22.84} & \underline{0.829} & \underline{0.167} 
    & \underline{22.87} & \underline{0.803} & \underline{0.167}
 \\
 
    Ours 
    & \textbf{21.15} & \textbf{0.737} & \textbf{0.201} 
    & \textbf{21.01} & \textbf{0.701} & \textbf{0.199} 
    & \textbf{26.42} & \textbf{0.897} & \textbf{0.104} 
    & \textbf{21.63} & \textbf{0.827} & \textbf{0.139} 
    & \textbf{26.21} & \textbf{0.907} & \textbf{0.102} 
    & \textbf{22.87} & \textbf{0.837} & \textbf{0.146} 
    & \textbf{23.22} & \textbf{0.818} & \textbf{0.149}
\\
    \bottomrule
    \end{tabular}}

%% file: images/qual_onthego.tex
\makebox[0.195\textwidth]{\footnotesize WildGaussians}
\makebox[0.195\textwidth]{\footnotesize SpotLessSplats}
\makebox[0.195\textwidth]{\footnotesize T-3DGS}
\makebox[0.195\textwidth]{\footnotesize Ours}
\makebox[0.195\textwidth]{\footnotesize GT}
\\
\includegraphics[width=0.195\textwidth]{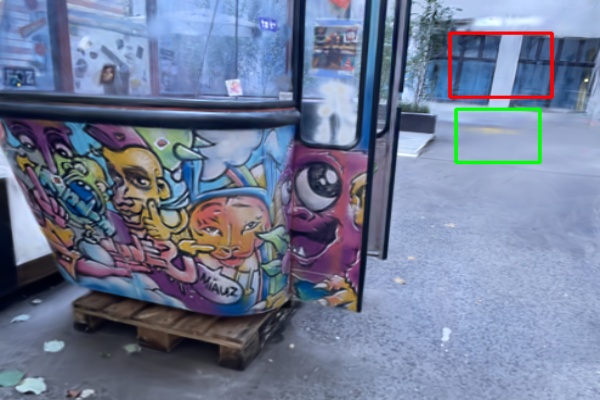}
\includegraphics[width=0.195\textwidth]{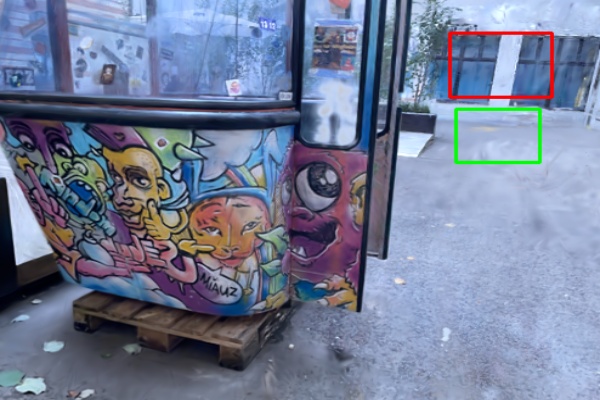}
\includegraphics[width=0.195\textwidth]{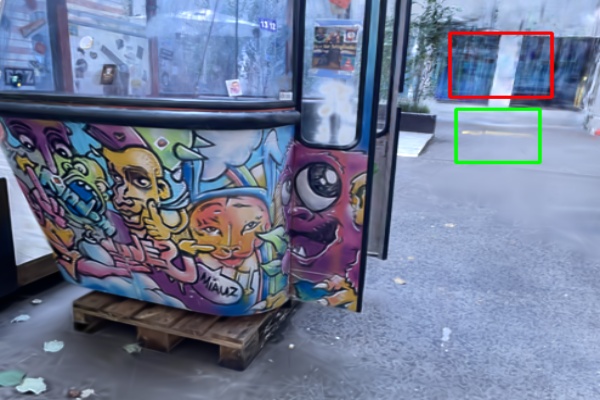}
\includegraphics[width=0.195\textwidth]{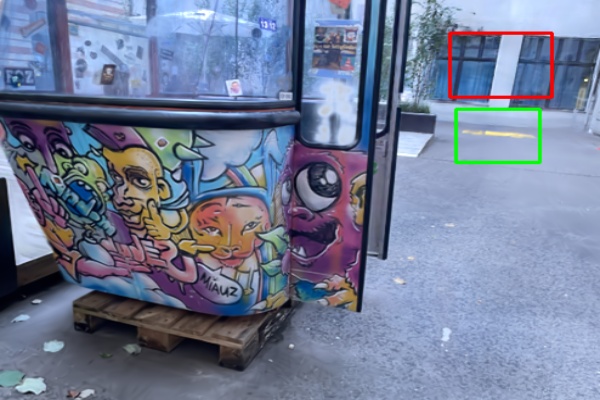}
\includegraphics[width=0.195\textwidth]{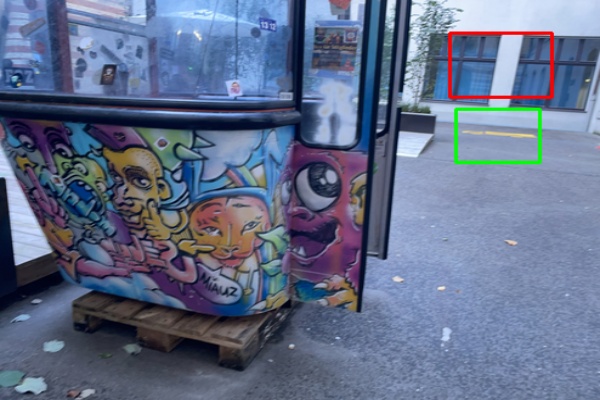}
\\
\includegraphics[width=0.095\textwidth,height=0.07\textwidth]{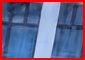}
\includegraphics[width=0.095\textwidth,height=0.07\textwidth]{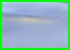}
\includegraphics[width=0.095\textwidth,height=0.07\textwidth]{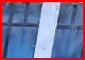}
\includegraphics[width=0.095\textwidth,height=0.07\textwidth]{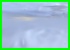}
\includegraphics[width=0.095\textwidth,height=0.07\textwidth]{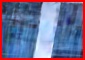}
\includegraphics[width=0.095\textwidth,height=0.07\textwidth]{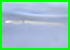}
\includegraphics[width=0.095\textwidth,height=0.07\textwidth]{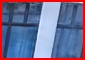}
\includegraphics[width=0.095\textwidth,height=0.07\textwidth]{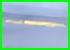}
\includegraphics[width=0.095\textwidth,height=0.07\textwidth]{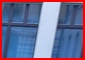}
\includegraphics[width=0.095\textwidth,height=0.07\textwidth]{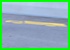}
\\
\includegraphics[width=0.195\textwidth]{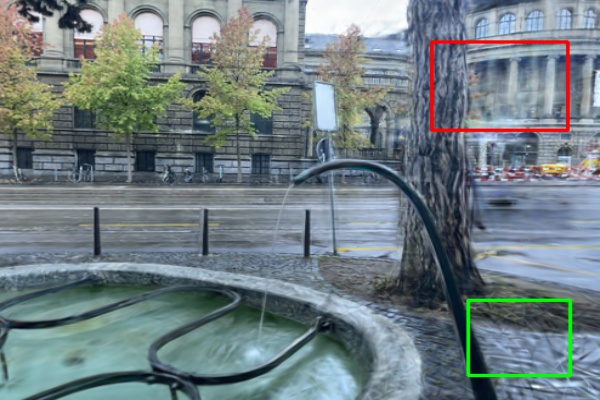}
\includegraphics[width=0.195\textwidth]{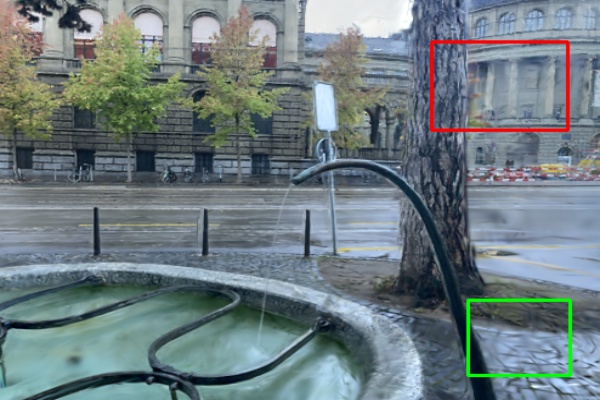}
\includegraphics[width=0.195\textwidth]{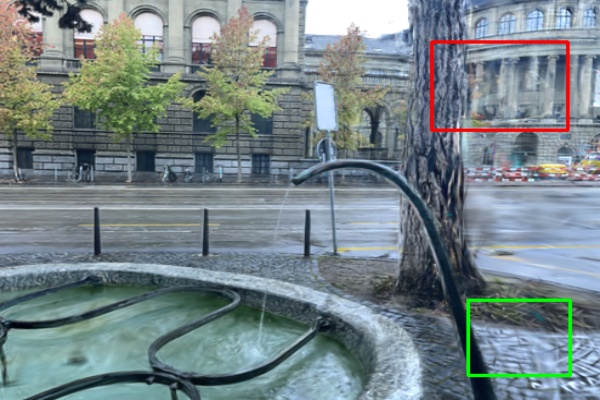}
\includegraphics[width=0.195\textwidth]{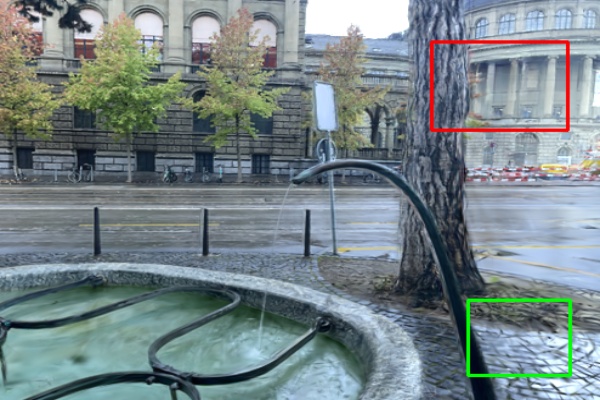}
\includegraphics[width=0.195\textwidth]{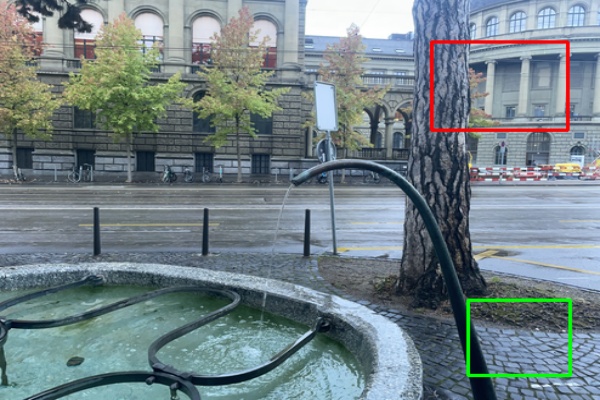}
\\
\includegraphics[width=0.095\textwidth,height=0.07\textwidth]{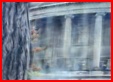}
\includegraphics[width=0.095\textwidth,height=0.07\textwidth]{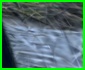}
\includegraphics[width=0.095\textwidth,height=0.07\textwidth]{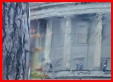}
\includegraphics[width=0.095\textwidth,height=0.07\textwidth]{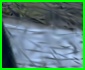}
\includegraphics[width=0.095\textwidth,height=0.07\textwidth]{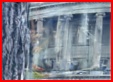}
\includegraphics[width=0.095\textwidth,height=0.07\textwidth]{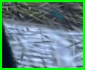}
\includegraphics[width=0.095\textwidth,height=0.07\textwidth]{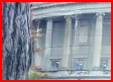}
\includegraphics[width=0.095\textwidth,height=0.07\textwidth]{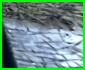}
\includegraphics[width=0.095\textwidth,height=0.07\textwidth]{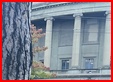}
\includegraphics[width=0.095\textwidth,height=0.07\textwidth]{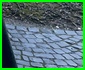}
\\

%% file: 5_experiments.tex
\section{Experiments}
\label{sec:Experiments}

\paragraph{Datasets} We evaluate our RobustSplat on two challenging benchmark datasets: \emph{NeRF On-the-go}~\cite{ren2024nerf} and \emph{RobustNeRF}~\cite{sabour2023robustnerf}. 
The NeRF On-the-go dataset consists of a total of 12 scenes, featuring varying occlusion levels ($5\% \sim 30\%$). Among them, six scenes are widely used, while the remaining six are more complex scenarios, referred to as \emph{NeRF On-the-go II} in this paper. 

We adopt the RobustNeRF dataset to further validate the effectiveness of the proposed method, which comprises four artificially designed indoor scenes, each incorporating various types of distractors that challenge reconstruction fidelity.

\paragraph{Implementation Details} Our codebase follows the official Gaussian Splatting (3DGS). During training, we adopt the same learning rate settings as 3DGS and set the total training iterations to 30K. The MLP consists of two linear layers, optimized with the Adam optimizer (learning rate set to 0.001). 
Fixed parameters are used for all experiments. The delayed iteration start is set to 10K, and the weights for MLP supervision terms are $\lambda_\textrm{residual}= 0.5$, $\lambda_\textrm{cos}= 0.5$, and $\lambda_\textrm{reg}= 2.0$, respectively. The mask regularization coefficient is $\beta_\textrm{reg}= 2000$. The features used by the MLP are extracted from DINOv2, with pre-trained weights from \emph{ViT-S/14 distilled}.

In the mask bootstrapping, the lowest spatial resolution features are extracted from images of size $(224\times224)$, while the highest spatial resolution features are derived from size $(504\times504)$. 
Following existing methods~\cite{sabour2024spotlesssplats}, we apply a downsampling factor of 8 on the NeRF On-the-go and RobustNeRF datasets (factor 4 for specific scenarios, e.g., arcdetriomphe and patio). Low-resolution residuals are further downsampled by an additional factor of 4 based on this configuration.

\paragraph{Baselines} We evaluated our RobustSplat against multiple baselines, including the vanilla 3D Gaussian Splatting~\cite{kerbl20233d} which we built upon, and recent robust methods including SpotLessSplats~\cite{sabour2024spotlesssplats}, WildGaussians~\cite{kulhanek2024wildgaussians}, Robust3DGaussians~\cite{ungermann2024robust} and T-3DGS~\cite{pryadilshchikov2024t}.
To ensure a fair comparison, we utilized the publicly available implementations of these methods and conducted evaluations using the same camera matrices across all experiments. 
We assessed performance through both visual comparisons and quantitative metrics, employing PSNR, SSIM, and LPIPS to measure reconstruction quality.

\subsection{Evaluation on \textbf{\emph{NeRF On-the-go}} Dataset}
\label{sub:Results I}
We first evaluate our method on NeRF On-the-go dataset. We can see from \Tref{table:onthego} that our method achieves best results across all six scenes on the PSNR, SSIM, and LPIPS metrics, clearly demonstrating the effectiveness of our method.

\Fref{fig:qual_onthego} shows the qualitative comparison, in which baseline approaches exhibit noticeable artifacts. Thanks to the proposed \delayedGS and \coarsetofine design, our method successfully eliminates these artifacts and achieves superior detail (e.g., the windows in \emph{Patio-high}, as well as the building in \emph{Fountain}).

\newcommand{\psnr}{\text{PSNR}}
\newcommand{\ssim}{\text{SSIM}}
\newcommand{\lpips}{\text{LPIPS}}

\subsection{Evaluation on the \textbf{\emph{RobustNeRF}} Dataset}
\label{sub:Results II}
To further validate the effectiveness of our method, we conduct comparisons with baseline methods on the RobustNeRF dataset, with quantitative results shown in \Tref{tab:robustnerf}. 
Our method achieves the best performance on the average metric.
Although our method performs slightly worse in PSNR and SSIM metrics on the \emph{Android} scene, it remains competitive with the state-of-the-art methods. 
In the remaining three scenes of the RobustNeRF dataset, our approach significantly outperforms existing methods. 
The qualitative comparisons are presented in \fref{fig:qual_robustnerf}, which shows that our method achieves transient-free reconstruction with sharp details.

\subsection{Ablation Study}
To evaluate the effectiveness of each component of our method, we built upon the 3DGS~\cite{kerbl20233d} and added different components to analyze the model performance.

\paragraph{Effects of Delayed Gaussian Growth} \Tref{table:ablation_method} shows that comparing with the full model, the model without \delayedGS (``3DGS+Mask+MB'') suffers from a noticeable decrease in all average metrics, which reiterate the effectiveness of the \delayedGS strategy in preventing the 3DGS to fit transient regions during the early optimization phase.

\begin{table}[t]
    \begin{center}
    \caption{Ablation of each component in our method on NeRF On-the-go datasets. ``3DGS+Mask'' is the model that integrate the transient mask estimation with 3DGS. We denote \emph{Mask Bootstrapping} as ``MB'', and \emph{\DelayedGS} as ``DG''. ``Full Model'' indicates ``3DGS+Mask+MB+DG''.}
    \label{table:ablation_method}
    \input{tables/ablation_method}
    \end{center}
    \vspace{-2em}
\end{table}

\begin{table*}[!t] \centering
    \begin{center}
    \captionof{table}{Quantitative results on RobustNeRF dataset~\cite{sabour2023robustnerf}. The best results are highlighted in \textbf{bold}, and the second in \underline{underline}.}
    \input{tables/robustnerf}
    \label{tab:robustnerf}
    \end{center}
    \vspace{-10pt}
    \input{images/qual_robustnerf}
    \captionof{figure}{Qualitative comparison on \emph{Crab2} and \emph{Statue} from RobustNeRF dataset}
    \label{fig:qual_robustnerf}
    \vspace{-1em}
\end{table*}
\paragraph{Effects of Mask Bootstrapping} 
\Tref{table:ablation_method} shows that removing the proposed \coarsetofine (``3DGS+Mask+DG'') leads to a decrease in overall performance. This drop is particularly evident in the Mountain scene, an unbounded environment with a large proportion of sky regions and sparsely initialized points, which results in overly smooth reconstructions during early optimization. Our mask bootstrapping provides more robust supervision, leading to more accurate reconstructions.

\paragraph{Effects of Mask Supervision} The supervisions of our mask MLP are derived from the image residuals and the feature similarities. 
We conduct experiments to evaluate the contribution of each component in helping identify transient objects.
As shown in \Tref{table:ablation_feat_sup}, removing either supervision component leads to varying degrees of degradation in metrics, indicating that their collaboration enables the more accurate transient region estimation, as shown in~\fref{fig:ablation_vis_mask}. 

\paragraph{Effects of Input Features for Mask Learning} 
Our method leverages DINOv2 features for mask prediction. 
We further investigate utilizing different feature representations as input to the mask MLP for mask prediction. 
As it is computationally intractable to extract the SD features for the rendered image at each iteration, we use the DINOv2 feature robust loss for training.  
As shown in \Tref{table:ablation_feat_sup}, the model trained with DINOv2 feature input achieves the best results. Note that our method can be seamlessly integrated with other feature representations.

\begin{table}[t] \centering
    \caption{Ablation of using different supervisions and input features for mask learning on NeRF On-the-go scenes.}
    \input{tables/ablation_feat_sup}

    \label{table:ablation_feat_sup}

    \vspace{0.3em}
    \makebox[0.115\textwidth]{\scriptsize GT Image}
    \makebox[0.115\textwidth]{\scriptsize Ours}
    \makebox[0.115\textwidth]{\scriptsize Ours w/o cosine}
    \makebox[0.115\textwidth]{\scriptsize Ours w/o residual}
    \\
    \includegraphics[width=0.48\textwidth]{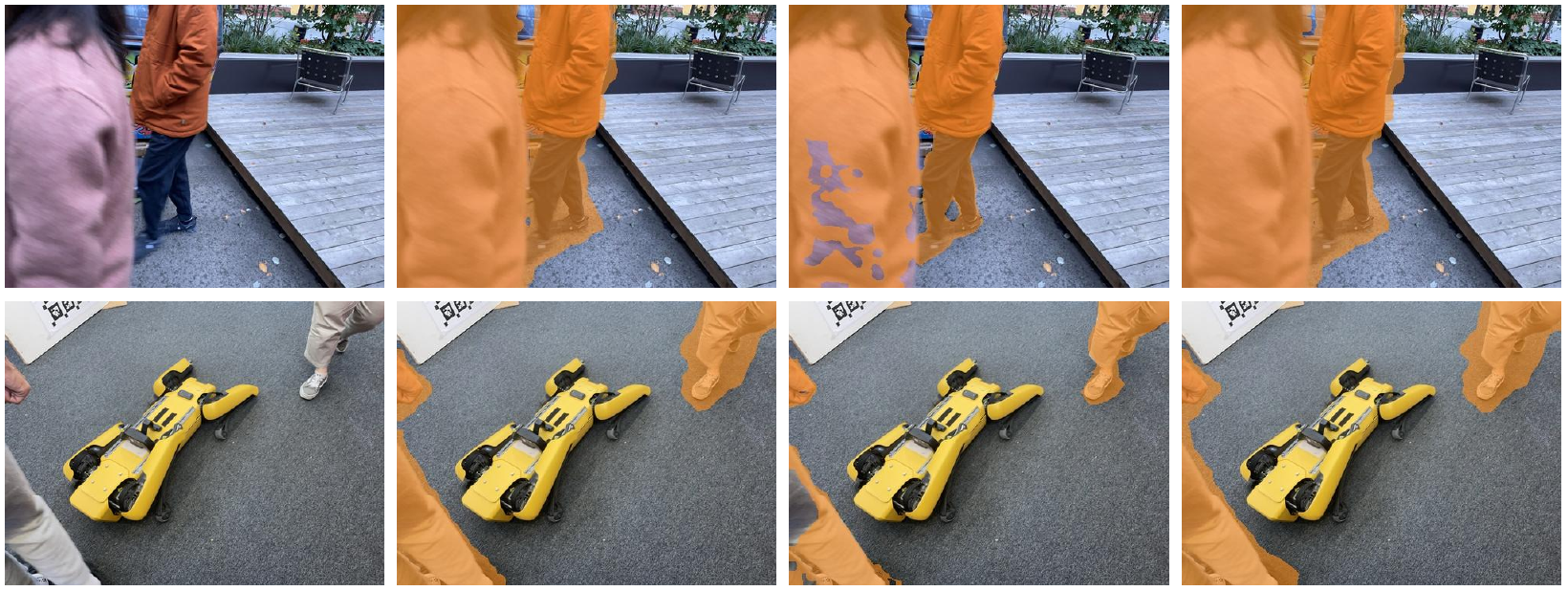}
    \captionof{figure}{Visualization of mask from different supervisions.} \label{fig:ablation_vis_mask}
    \vspace{-1em}
\end{table}

%% file: tables/ablation_method.tex
\resizebox{0.48\textwidth}{!}{
    \begin{tabular}{l|cc|cc|cc|cc|cc|cc}
    \toprule
    \multirow{2}{*}{Method} 
    & \multicolumn{2}{c|}{Mountain}  
    & \multicolumn{2}{c|}{Fountain} 
    & \multicolumn{2}{c|}{Corner}  
    & \multicolumn{2}{c|}{Patio} 
    & \multicolumn{2}{c|}{Spot} 
    & \multicolumn{2}{c}{Patio-High} \\
    & PSNR & SSIM 
    & PSNR & SSIM 
    & PSNR & SSIM 
    & PSNR & SSIM 
    & PSNR & SSIM 
    & PSNR & SSIM \\
    \midrule
    3DGS~\cite{kerbl20233d}
    & 19.21 & 0.691
    & 20.08 & 0.686
    & 22.65 & 0.835 
    & 17.04 & 0.713
    & 18.54 & 0.717
    & 17.04 & 0.657\\
    \phantom{00}+ Mask
    & 19.81 & 0.701 
    & 20.74 & 0.691 
    & 25.05 & 0.877 
    & 21.23 & 0.820 
    & 24.75 & 0.903 
    & 22.19 & 0.832 \\
    \phantom{0000}+ DG
    & 20.85 & 0.721 
    & 20.99 & 0.701 
    & 26.01 & 0.896 
    & 21.49 & 0.827 
    & 25.61 & 0.906 
    & 22.74 & \textbf{0.838} \\
    \phantom{0000}+ MB
    & 20.78 & 0.713 
    & 20.83 & 0.692 
    & 25.52 & 0.885 
    & 20.88 & 0.817 
    & 25.25 & 0.900 
    & 22.11 & 0.826 \\
    Full Model 
    & \textbf{21.15} & \textbf{0.737}
    & \textbf{21.01} & \textbf{0.701}
    & \textbf{26.42} & \textbf{0.897}
    & \textbf{21.63} & \textbf{0.827}
    & \textbf{26.21} & \textbf{0.907}
    & \textbf{22.87} & 0.837\\
    \bottomrule
    \end{tabular}}

%% file: tables/robustnerf.tex
\resizebox{\textwidth}{!}{
\begin{tabular}{l|*{3}{c}|*{3}{c}|*{3}{c}|*{3}{c}|*{3}{c}}
    \toprule
    & \multicolumn{3}{c|}{Android} 
    & \multicolumn{3}{c|}{Crab2}
    & \multicolumn{3}{c|}{Statue}
    & \multicolumn{3}{c|}{Yoda}
    & \multicolumn{3}{c}{Mean}
    \\
    Method
    & \multicolumn{1}{c}{PSNR} 
    & \multicolumn{1}{c}{SSIM} 
    & \multicolumn{1}{c|}{LPIPS} 
    & \multicolumn{1}{c}{PSNR} 
    & \multicolumn{1}{c}{SSIM} 
    & \multicolumn{1}{c|}{LPIPS} 
    & \multicolumn{1}{c}{PSNR} 
    & \multicolumn{1}{c}{SSIM} 
    & \multicolumn{1}{c|}{LPIPS} 
    & \multicolumn{1}{c}{PSNR} 
    & \multicolumn{1}{c}{SSIM} 
    & \multicolumn{1}{c|}{LPIPS} 
    & \multicolumn{1}{c}{PSNR} 
    & \multicolumn{1}{c}{SSIM} 
    & \multicolumn{1}{c}{LPIPS} 
    \\
    \midrule
    3DGS~\cite{kerbl20233d}
    & 23.32 & 0.794 & 0.159 
    & 31.76 & 0.925 & 0.172 
    & 20.83 & 0.830 & 0.148 
    & 28.92 & 0.905 & 0.192
    & 26.21 & 0.864 & 0.168
    \\
    SpotLessSplats~\cite{sabour2024spotlesssplats}
    & 24.20 & 0.810 & 0.159 
    & \underline{33.90} & \underline{0.933} & 0.169 
    & 21.97 & 0.821 & 0.163 
    & \underline{34.24} & \underline{0.938} & \underline{0.156} 
    & \underline{28.58} & 0.875 & 0.162
    \\
    WildGaussians~\cite{kulhanek2024wildgaussians}
    & \underline{24.67} & 0.828 & 0.151 
    & 30.52 & 0.909 & 0.213 
    & \underline{22.54} & 0.863 & 0.129 
    & 30.55 & 0.905 & 0.202
    & 27.07 & 0.876 & 0.174
    \\
    Robust3DGaussians~\cite{ungermann2024robust}
    & 24.30 & 0.813 & 0.134 
    & 32.77 & 0.926 & \underline{0.162} 
    & 21.93 & 0.837 & 0.135 
    & 30.85 & 0.913 & 0.177 
    & 27.46 & 0.872 & {0.152}
    \\
    T-3DGS~\cite{pryadilshchikov2024t}
    & \textbf{24.81} & \textbf{0.839} & \textbf{0.125} 
    & 32.97 & 0.929 & 0.177 
    & 22.53 & \underline{0.864} & \underline{0.113} 
    & 32.68 & 0.920 & 0.182 
    & 28.25 & \underline{0.888} & \underline{0.149}
    
    \\
    Ours
    & 24.62 & \underline{0.831} & \textbf{0.125} 
    & \textbf{34.88} & \textbf{0.940} & \textbf{0.154} 
    & \textbf{22.80} & \textbf{0.865} & \textbf{0.110} 
    & \textbf{35.14} & \textbf{0.944} & \textbf{0.151} 
    & \textbf{29.36} & \textbf{0.895} & \textbf{0.135}
    \\
    \bottomrule
\end{tabular}
}

%% file: images/qual_robustnerf.tex
\makebox[0.195\textwidth]{\footnotesize WildGaussians}
\makebox[0.195\textwidth]{\footnotesize SpotLessSplats}
\makebox[0.195\textwidth]{\footnotesize T-3DGS}
\makebox[0.195\textwidth]{\footnotesize Ours}
\makebox[0.195\textwidth]{\footnotesize GT}
\\
\includegraphics[width=0.195\textwidth]{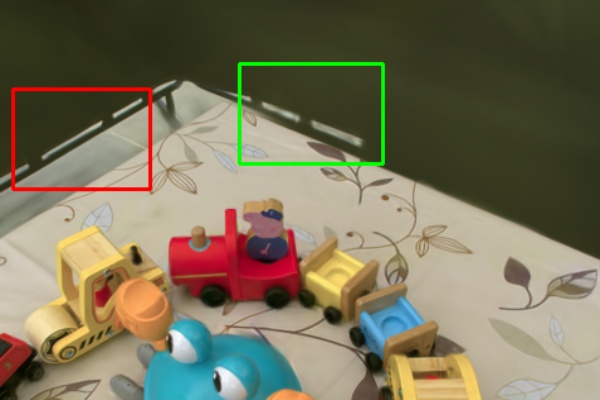}
\includegraphics[width=0.195\textwidth]{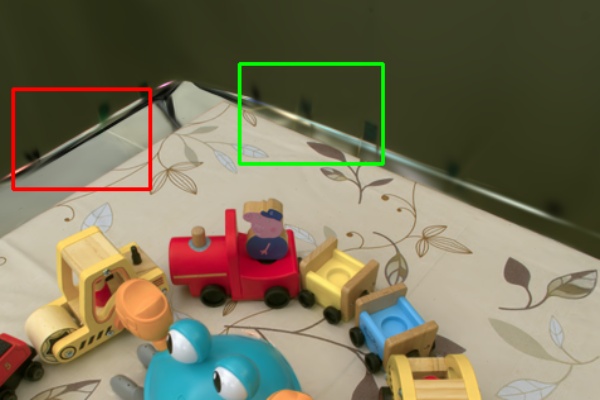}
\includegraphics[width=0.195\textwidth]{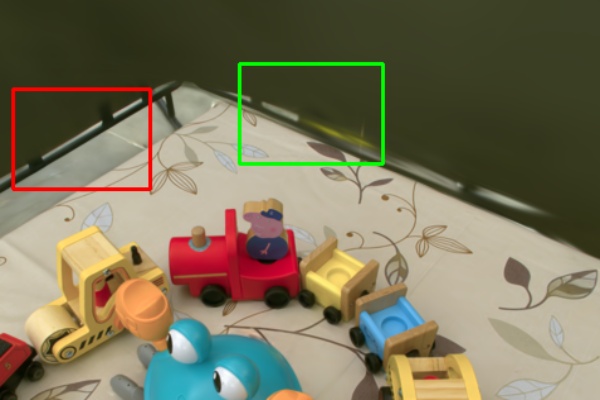}
\includegraphics[width=0.195\textwidth]{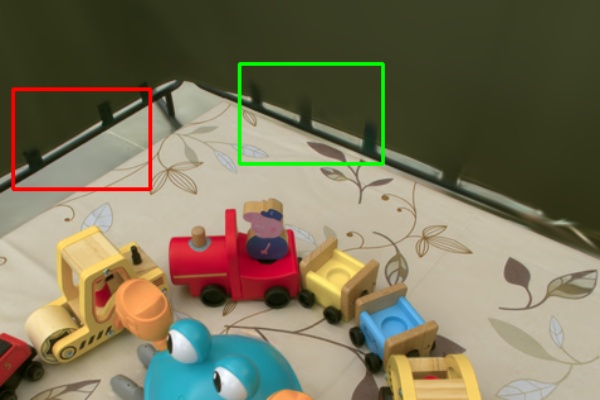}
\includegraphics[width=0.195\textwidth]{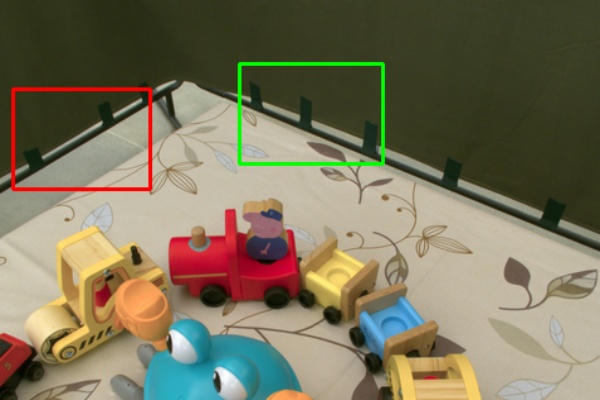}
\\
\includegraphics[width=0.095\textwidth,height=0.07\textwidth]{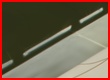}
\includegraphics[width=0.095\textwidth,height=0.07\textwidth]{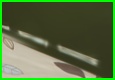}
\includegraphics[width=0.095\textwidth,height=0.07\textwidth]{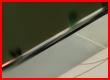}
\includegraphics[width=0.095\textwidth,height=0.07\textwidth]{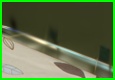}
\includegraphics[width=0.095\textwidth,height=0.07\textwidth]{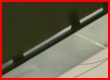}
\includegraphics[width=0.095\textwidth,height=0.07\textwidth]{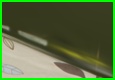}
\includegraphics[width=0.095\textwidth,height=0.07\textwidth]{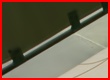}
\includegraphics[width=0.095\textwidth,height=0.07\textwidth]{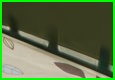}
\includegraphics[width=0.095\textwidth,height=0.07\textwidth]{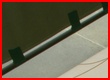}
\includegraphics[width=0.095\textwidth,height=0.07\textwidth]{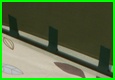}
\\\includegraphics[width=0.195\textwidth]{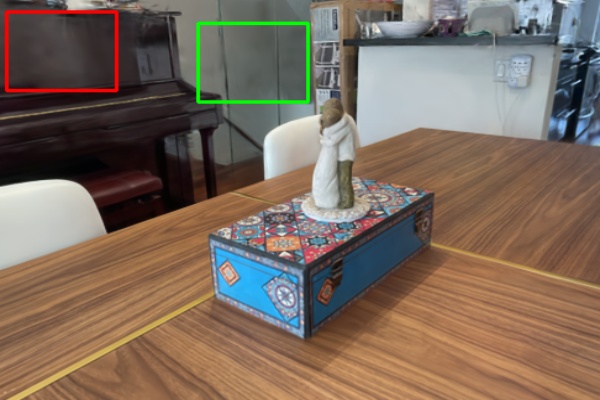}
\includegraphics[width=0.195\textwidth]{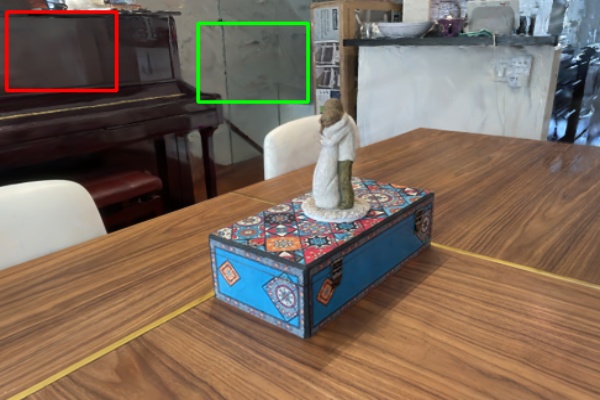}
\includegraphics[width=0.195\textwidth]{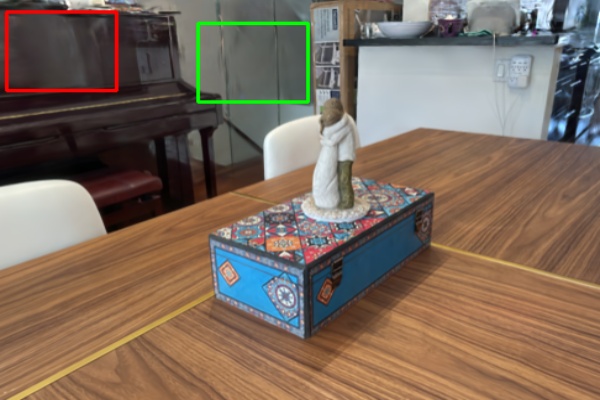}
\includegraphics[width=0.195\textwidth]{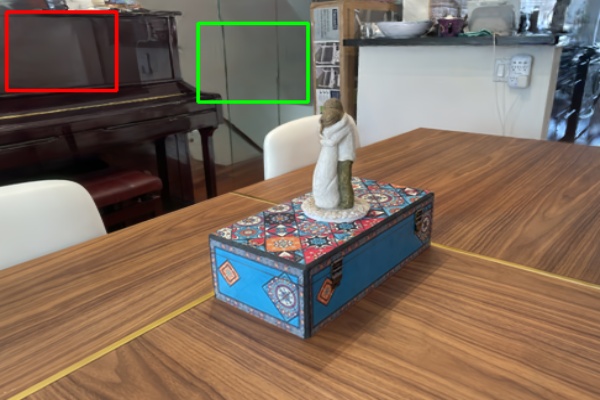}
\includegraphics[width=0.195\textwidth]{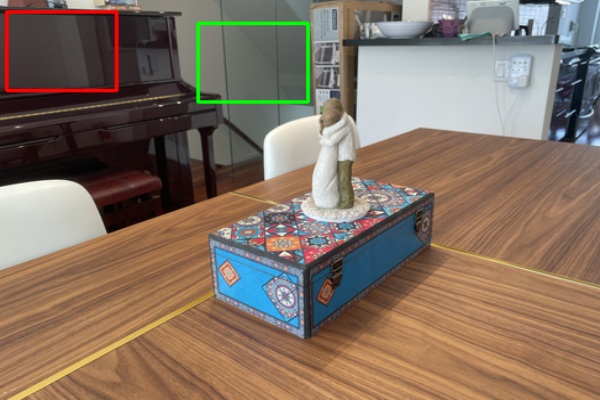}
\\
\includegraphics[width=0.095\textwidth,height=0.07\textwidth]{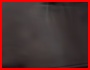}
\includegraphics[width=0.095\textwidth,height=0.07\textwidth]{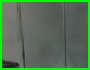}
\includegraphics[width=0.095\textwidth,height=0.07\textwidth]{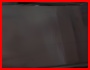}
\includegraphics[width=0.095\textwidth,height=0.07\textwidth]{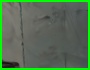}
\includegraphics[width=0.095\textwidth,height=0.07\textwidth]{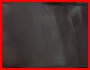}
\includegraphics[width=0.095\textwidth,height=0.07\textwidth]{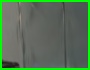}
\includegraphics[width=0.095\textwidth,height=0.07\textwidth]{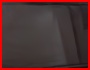}
\includegraphics[width=0.095\textwidth,height=0.07\textwidth]{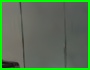}
\includegraphics[width=0.095\textwidth,height=0.07\textwidth]{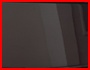}
\includegraphics[width=0.095\textwidth,height=0.07\textwidth]{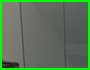}
\\

%% file: tables/ablation_feat_sup.tex
\resizebox{0.48\textwidth}{!}{
    \begin{tabular}{l|ccc|ccc|ccc}
    \toprule
    \multirow{2}{*}{Method}  & \multicolumn{3}{c|}{Spot} & \multicolumn{3}{c|}{Patio-High} & \multicolumn{3}{c}{Mean} \\
    & \psnr & \ssim & \lpips & \psnr & \ssim & \lpips & \psnr & \ssim & \lpips \\
    \midrule
    Ours w/o residual & 25.24 & 0.900 & 0.110 & 22.29 & 0.821 & 0.154 & 23.77 & 0.861 & 0.132 \\ 
    Ours w/o cosine & 24.29 & 0.894 & 0.113 & 22.62 & 0.830 & 0.154 & 23.46 & 0.862 & 0.134 \\
    Ours
    & \textbf{26.21} & \textbf{0.907} & \textbf{0.102} 
    & \textbf{22.87} & \textbf{0.837} & \textbf{0.146} 
    & \textbf{24.54} & \textbf{0.872} & \textbf{0.124}\\
    \midrule
    Ours w/ SAM2 
    & 25.31 & 0.907 & \textbf{0.101} 
    & 22.70 & \textbf{0.838} & 0.145 
    & 24.01 & \textbf{0.873} & \textbf{0.123} \\
    Ours w/ SD 
    & 25.35 & 0.904 & 0.103 
    & 22.73 & 0.836 & \textbf{0.145} 
    & 24.04 & 0.870 & 0.124 \\
    Ours
    & \textbf{26.21} & \textbf{0.907} & 0.102 
    & \textbf{22.87} & 0.837 & 0.146 
    & \textbf{24.54} & 0.872 & 0.124\\
    \bottomrule
    \end{tabular}}

%% file: 6_conclusions.tex
\section{Conclusion}
\label{sec:Conclusion}

In this work, we introduce RobustSplat, a robust framework for transient-free 3D Gaussian Splatting, effectively mitigating artifacts caused by transient objects in dynamic scenes. 
Built on our analysis on the relation between Gaussians densification and artifacts caused by transient objects, our approach integrates a delayed Gaussian growth strategy to prioritize static scene optimization and a scale-cascaded mask bootstrapping method for reliable transient object suppression. Through comprehensive experiments on multiple challenging datasets, RobustSplat demonstrates superior robustness and rendering quality compared to existing methods.

\paragraph{Limitations} 
Our current approach focuses solely on transient object removal without explicitly handling illumination changes, which limits the applicability of our method in more uncontrolled environments. In future work, we aim to investigate illumination-aware solutions to model lighting changes by incorporating the characteristics of the Gaussian densification process.

%% file: 7_acknowledgements.tex
\section*{Acknowledgements}
This work was in part supported by the National Key $\textrm{R\&D}$ Program of China (Grant No.~2022ZD0119200), NSFC (Grant Nos.~62202409, 62472453), Guangdong Natural Science Foundation (No.~2025A1515010782), Shenzhen Science and Technology Program (No.~JCYJ20220818102012025), CIE-Smartchip research fund (No.~2024-08), the Key Technology Project of Shenzhen (Grant No.~KJZD20230923115104009), and Guangdong Provincial Key Laboratory of Future Networks of Intelligence (Grant No.~2022B1212010001).

%% file: 8_supp.tex
\def\maketitlesupplementarysinglecolumn
{
    \newpage
    \begin{center}
        {\Large
        \textbf{\thetitle}\\
        \vspace{0.5em}Supplementary Material \\
        \vspace{1.0em}
        }
    \end{center}
}
\onecolumn
\setcounter{figure}{0}
\setcounter{table}{0}
\setcounter{page}{1}
\maketitlesupplementarysinglecolumn
\renewcommand{\thetable}{S\arabic{table}}
\renewcommand{\thefigure}{S\arabic{figure}}

\section{Discussions}

\paragraph{Sparse Gaussian Initialization and Gaussian Densification}
The optimization of 3D Gaussian Splatting (3DGS) relies on an initial set of points obtained via Structure-from-Motion (SfM). Since SfM reconstructs sparse point clouds based on multi-view consistency, transient objects that remain stationary in multiple captured images before moving can introduce noisy points into the reconstruction. As a result, 3DGS may initially fit these transient regions, even before Gaussian densification takes place.

\begin{figure*}[h] \centering
    \includegraphics[width=0.7\textwidth]{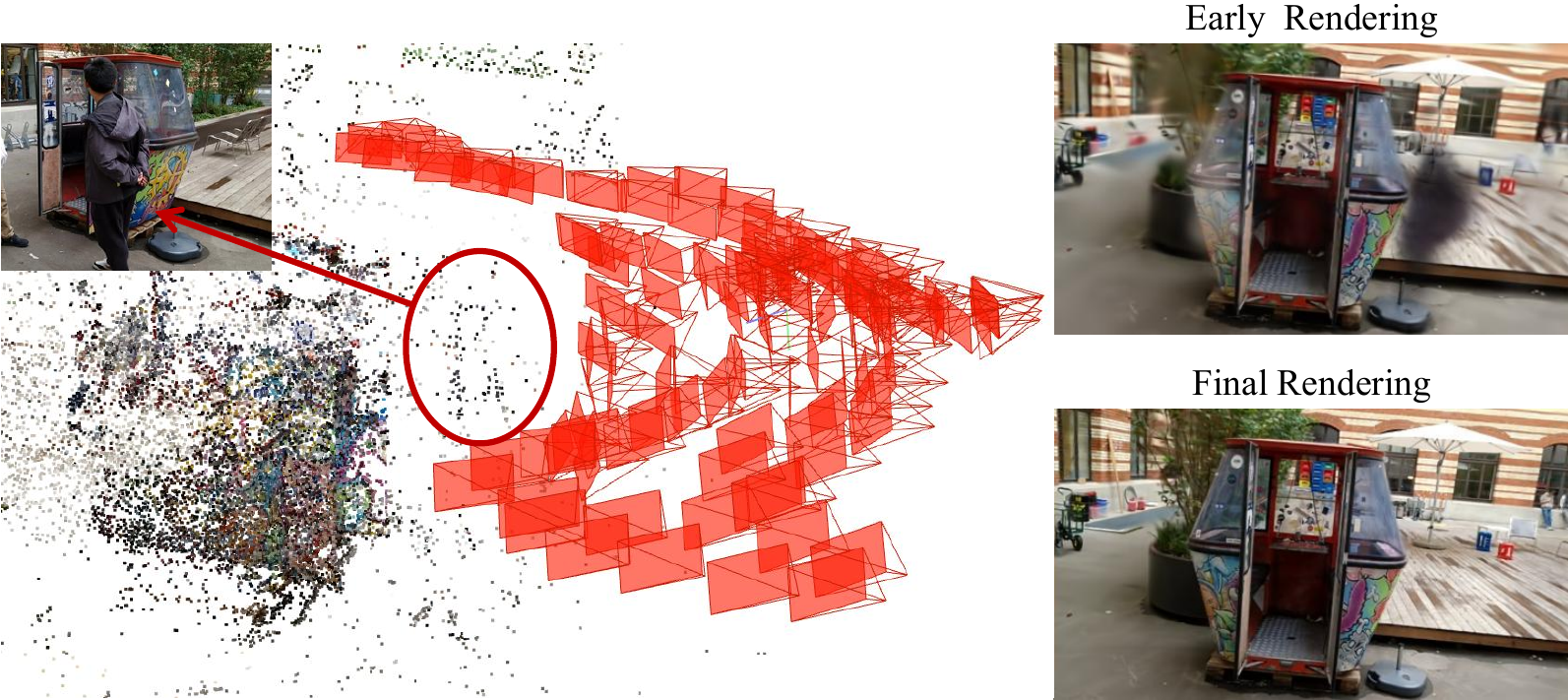}
    \caption{Gaussians initialization with inaccurate COLMAP SfM point clouds may affect the early optimization stage.} \label{fig:noisy_sparse}
\end{figure*}

As illustrated in \fref{fig:noisy_sparse}, in the \emph{Patio} scene from the NeRF On-the-go dataset, moving subjects remained stationary for a period, leading to COLMAP reconstructing noisy points corresponding to these transient objects. As a result, 3DGS initially fits to these transient regions. However, with longer optimization, our transient mask estimation progressively removes these artifacts.
This observation highlights that by applying a transient mask to filter dynamic regions, our method effectively mitigates the impact of noisy initialization, leading to improved reconstruction quality.

\begin{figure*}[h] \centering
    \makebox[0.30\textwidth]{\small GT Image}
    \makebox[0.30\textwidth]{\small Rendering}\\
    \includegraphics[width=0.6\textwidth]
    {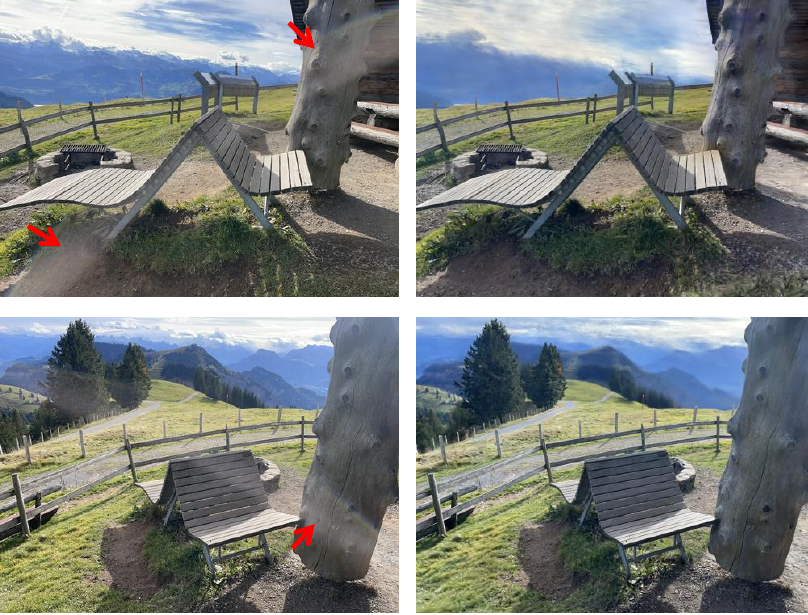}
    \caption{Illumination changes in real-world scenes.} \label{fig:illumination}
\end{figure*}

\paragraph{Illumination Variations}
In real-world environments, besides transient disturbances, illumination changes can introduce multi-view inconsistencies, leading to floating artifacts. Our method mainly addresses transient object interference. However, when abrupt illumination changes occur in a scene, our approach fails to correctly model the actual lighting variations due to the absence of an explicit illumination model~\fref{fig:illumination}. A promising direction for future work is to incorporate illumination modeling into our method, enabling the handling of more complex outdoor datasets.

\paragraph{Feature Extraction for Mask Estimation}
In the main text, we discuss the impact of different feature types on mask learning. DINOv2 performs well due to its efficiency and the reliable consistency of features within similar object categories. However, its patch-based nature introduces inconsistencies at the edges when extended to high-resolution settings, limiting the effectiveness of our mask predictor.
In this work, we slightly expand the mask by applying dilation with a kernel size of 7.
In the future, we will explore integrating more expressive and efficient feature extractors for mask learning.

\section{More Details for the Method}
\paragraph{Training Details}
The original 3DGS ~\cite{kerbl20233d} resets the opacity starting from the 3000 iterations while maintaining an interval of 3000 iterations. This operation aims to eliminate the accumulation of low-opacity Gaussian primitives in regions close to the camera, which can interfere with gradient backpropagation and manifest as artifacts.
However, the opacity reset is no longer suitable for our method due to the delayed Gaussian growth. Therefore, we delay the opacity reset to start from the 15000 iterations while maintaining the same interval of 3000 iterations. Meanwhile, the start of pruning is also delayed to 10000 iterations to align with delayed Gaussian growth.

\paragraph{Robust Loss based on Image Residuals}
The image robust loss used in our mask predictor follows ~\cite{sabour2024spotlesssplats}:
\begin{equation}
    \mathcal{L}_\textrm{residual} = max\left(\left( \textrm{U}-\textrm{M} \right),0 \right) + max\left(\left( \textrm{M}-\textrm{L} \right),0 \right),
\end{equation}
where $\textrm{M}$ is the mask we predicted, $\textrm{U}$ and $\textrm{L}$ are upper and lower bound of the dynamic residual mask, respectively, which determined by different values of the parameter $\tau$. 
In our method, the parameters are set to $\tau_{u}=0.6$ and $\tau_{l}=0.8$ for all experiments.

\section{Runtime Evaluation}
Our method adopts the lightweight DINOv2 model \textit{ViT‑S/14‑distilled}, with a feature dimensionality of 384, for feature extraction. 
As shown in \Tref{table:time_comparison}, our method runs slightly slower than 3DGS but remains faster than other methods. 
SpotLessSplats achieves similar optimization time without iterative feature extraction, but its SD features, with a dimensionality of 1280, require a long processing time before training.  
\begin{table}[h]
    \vspace{-0.2em}
    \begin{center}
    \caption{Runtime evaluation on an NVIDIA RTX 3090 (unit: minutes). The runtime of SpotLessSplats is divided into two parts: training and SD feature extraction.}
    \label{table:time_comparison}
    \input{rebuttal/table_time}
    \end{center}
    \vspace{-2.0em}
\end{table}

\section{More Ablation Study}

\paragraph{Effects of Mask Regularization.}
Initial mask estimation yields suboptimal results in most scenes due to unconverged reconstruction at early training stages. To address this challenge, we introduce a mask regularization for stabilizing early-stage mask training. \Tref{table:maskreg} shows that removing the proposed mask regularization leads to a decrease in overall performance.
\begin{table}[h]
    \begin{center}
    \caption{Effects of Mask Regularization. We denote \textit{Mask Regularization} as ``MR''.}
    \label{table:maskreg}
    \input{rebuttal/table_maskreg}
    \end{center}
    \vspace{-1.0em}
\end{table}

\paragraph{Effects of Delayed Gaussian Growth.} 
We discussed the effectiveness of Delayed Gaussian Growth in Section 4.3 of the main paper. To further validate its effects, we extend the Delayed Gaussian Growth to 3DGS in this supplementary material. \Tref{table:3dgs_dg} shows that integrating the delayed Gaussian growth into 3DGS leads to improve results, but its performance is limited by the lack of predicting the transient masks.
\begin{table}[t]
    \begin{center}
    \caption{Effects of Extending Delayed Gaussian Growth to 3DGS. We denote \textit{Delayed Gaussian Growth} as ``DG''.}
    \label{table:3dgs_dg}
    \input{rebuttal/table_3dgsdg}
    \end{center}
    \vspace{-2em}
\end{table}

\section{Evaluation on On-the-go II Dataset}
The NeRF On-the-go II dataset~\cite{ren2024nerf} is more challenging compared to the other scenes of NeRF On-the-go, as it consists of outdoor scenes that include not only dynamic objects but also motion blur and varying lighting conditions. Since the testing images in the On-the-go II dataset contain moving objects, we manually segment and exclude these objects when computing the metrics to ensure a fair evaluation.

We can see from \Tref{table:onthego2} that our method achieves nearly the best results across all six scenes, except for the second-best performance in the PSNR metric on \emph{Statue}. Moreover, our method outperforms existing methods and achieves state-of-the-art regarding average metrics.
\Fref{fig:vis_onthego2_1}, \Fref{fig:vis_onthego2_2}, and \Fref{fig:vis_onthego2_3} present qualitative comparisons with existing methods on the NeRF On-the-go II dataset. Our method successfully eliminates artifacts (\eg, vehicles in the \emph{Drone}) and recovers finer details (\eg, thin cables in the \emph{Train-station}), further demonstrating its effectiveness in handling complex scenarios.

\begin{table*}[h] \centering
    \begin{center}
    \captionof{table}{Quantitative comparison on NeRF On-the-go II Dataset. The best results are highlighted in \textbf{bold}, and the second in \underline{underline}.}
    \label{table:onthego2}
    \input{supp/onthego2}
    \end{center}
    \vspace{-2em}
\end{table*}

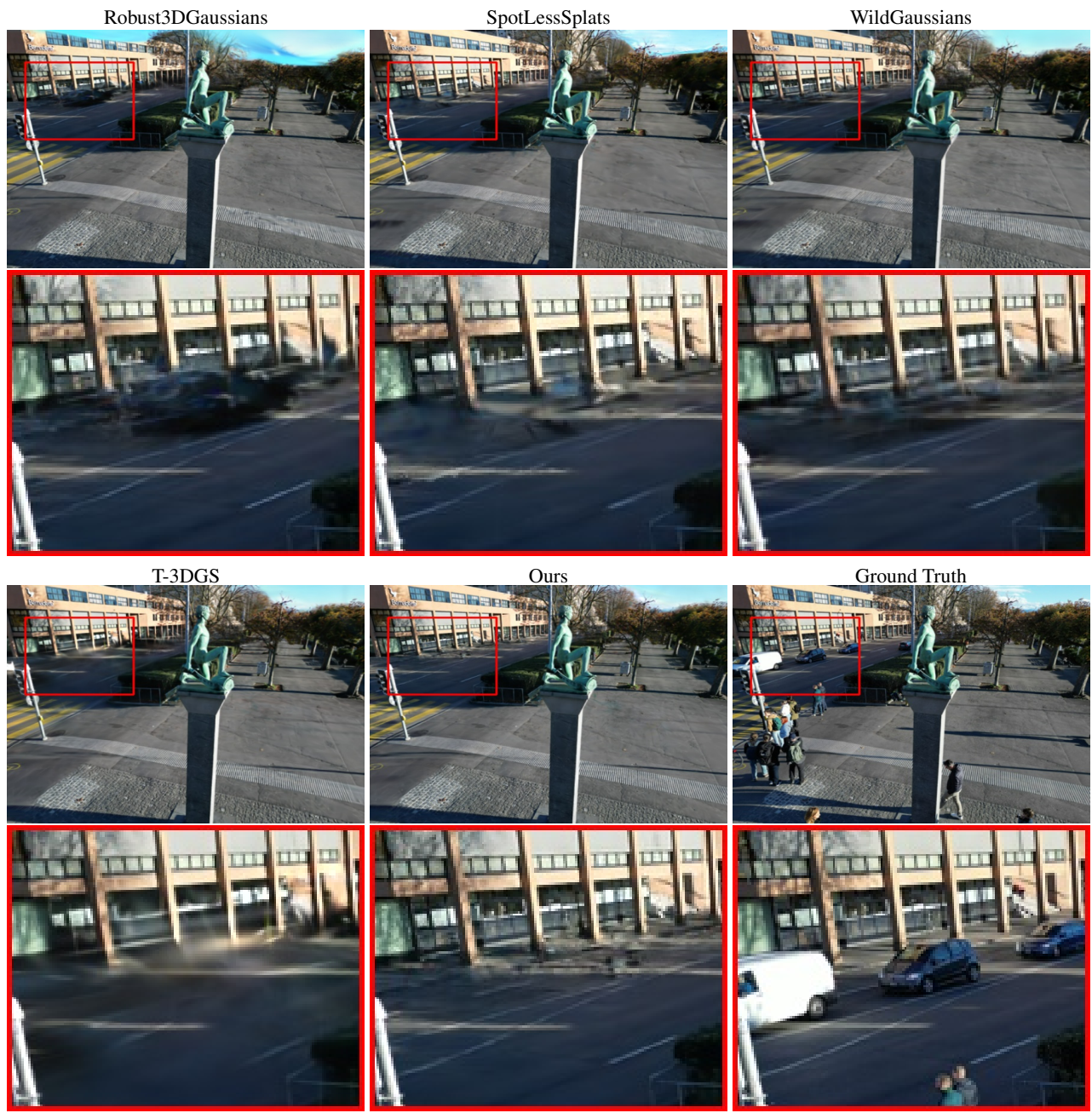
\begin{figure*}[h] \centering
    \input{supp/vis_onthego2_1}
    \caption{Qualitative results on \emph{Drone} in NeRF On-the-go II dataset.} \label{fig:vis_onthego2_1}
\end{figure*}
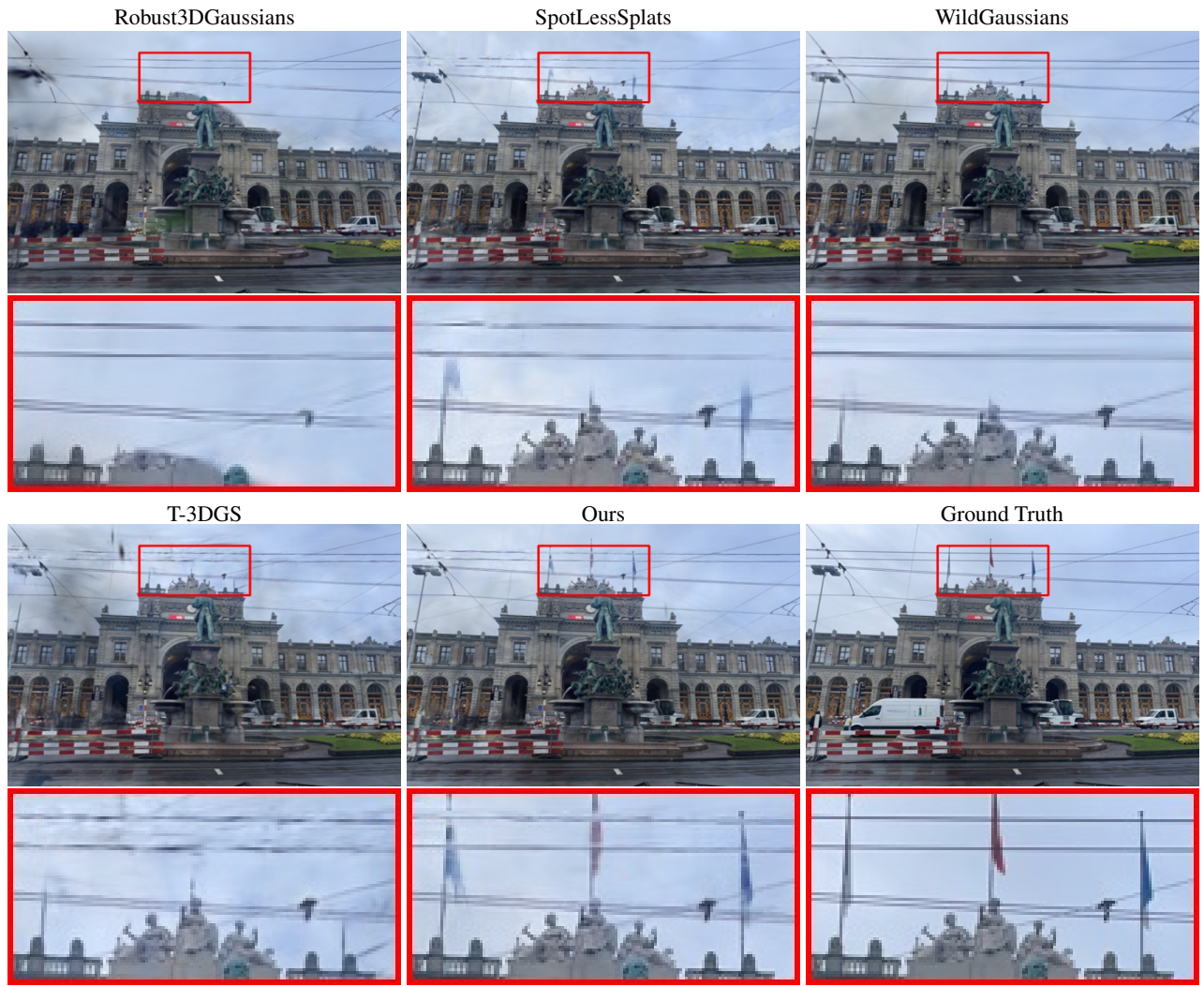
\begin{figure*}[h] \centering
    \input{supp/vis_onthego2_2}
    \caption{Qualitative results on \emph{Train-station} in NeRF On-the-go II dataset.} \label{fig:vis_onthego2_2}
\end{figure*}
\begin{figure*}[h] \centering
    \input{supp/vis_onthego2_3}
    \caption{Qualitative results on \emph{Tree} in NeRF On-the-go II dataset.} \label{fig:vis_onthego2_3}
\end{figure*}

\section{Comparison of Mask Estimation}
\begin{figure*}[h] \centering
    \makebox[0.231\textwidth]{\normalsize SpotlessSplats}
    \makebox[0.231\textwidth]{\normalsize Robust3DGS}
    \makebox[0.231\textwidth]{\normalsize T-3DGS}
    \makebox[0.231\textwidth]{\normalsize Ours}\\
    \includegraphics[width=0.95\textwidth]
    {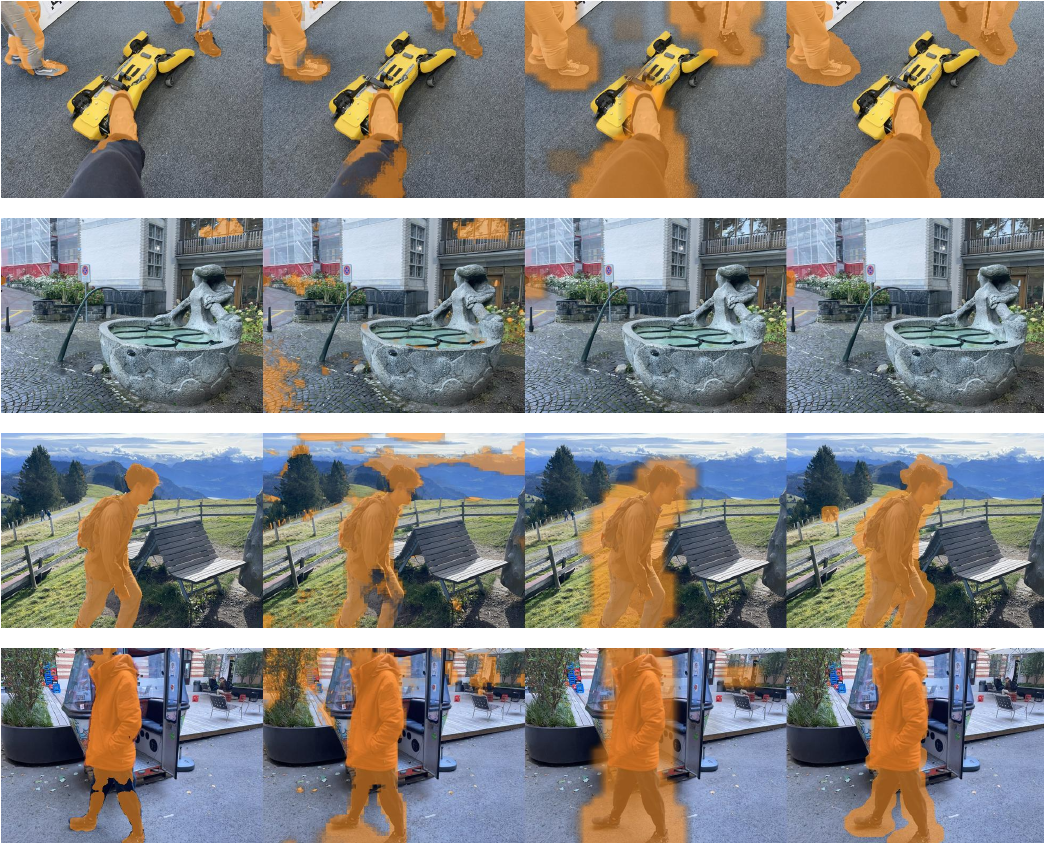}
    \caption{Comparison of transient mask in NeRF On-the-go dataset.} \label{fig:mask_estimation}
\end{figure*}

\Fref{fig:mask_estimation} compares the transient mask estimation results of our method with existing methods. Our method can better filter the transient objects while keeping the static regions, leading to less artifacts and sharp details in the rendering images.

%% file: rebuttal/table_time.tex
\resizebox{0.68\textwidth}{!}{
    \begin{tabular}{lcccccc}
    \toprule
    \multirow{2}{*}{Method} & Mountain & Fountain & Corner & Patio & Spot & Patio-High \\
     & \#Img 120 & \#Img 169 & \#Img 101 & \#Img 99 & \#Img 169 & \#Img 222 \\
    \midrule
    3DGS & 12.21 & 14.37 & 9.986 & 7.707 & 11.68 & 12.82 \\
    SpotLessSplats & 13.48+6.9 & 16.07+9.8 & 14.15+6.4 & 13.82+6.4 & 13.03+9.5 & 14.07+13.7 \\
    WildGaussians & 32.63 & 52.90 & 33.58 & 29.93 & 27.32 & 33.86 \\
    Ours & 15.43 & 17.33 & 13.32 & 12.82 & 12.95 & 14.35\\
    \bottomrule
    \end{tabular}
}

%% file: rebuttal/table_maskreg.tex
\resizebox{0.82\textwidth}{!}{
    \begin{tabular}{l|cc|cc|cc|cc|cc|cc}
    \toprule
    \multirow{2}{*}{Method} & \multicolumn{2}{c|}{Mountain} & \multicolumn{2}{c|}{Fountain} & \multicolumn{2}{c|}{Corner} & \multicolumn{2}{c|}{Patio} & \multicolumn{2}{c|}{Spot} & \multicolumn{2}{c}{Patio-high} \\
     & PSNR & SSIM & PSNR & SSIM & PSNR & SSIM & PSNR & SSIM & PSNR & SSIM & PSNR & SSIM \\
    \midrule
    Ours w/o MR & 21.09 & 0.728 & 20.87 & \textbf{0.701} & 26.18 & 0.889 & 21.61 & 0.826 & 25.63 & \textbf{0.907} & 22.68 & \textbf{0.837}\\
    Ours & \textbf{21.15} & \textbf{0.737} & \textbf{21.01} & \textbf{0.701} & \textbf{26.42} & \textbf{0.897} & \textbf{21.63} & \textbf{0.827} & \textbf{26.21} & \textbf{0.907} & \textbf{22.87} & \textbf{0.837} \\
    \bottomrule
    \end{tabular}
}

%% file: rebuttal/table_3dgsdg.tex
\resizebox{0.82\textwidth}{!}{
    \begin{tabular}{l|cc|cc|cc|cc|cc|cc}
    \toprule
    \multirow{2}{*}{Method} & \multicolumn{2}{c|}{Mountain} & \multicolumn{2}{c|}{Fountain} & \multicolumn{2}{c|}{Corner} & \multicolumn{2}{c|}{Patio} & \multicolumn{2}{c|}{Spot} & \multicolumn{2}{c}{Patio-high} \\
     & PSNR & SSIM & PSNR & SSIM & PSNR & SSIM & PSNR & SSIM & PSNR & SSIM & PSNR & SSIM \\
    \midrule
    3DGS & 19.21 & 0.691 & 20.08 & 0.686 & 22.65 & 0.835 & 17.04 & 0.713 & 18.54 & 0.717 & 17.04 & 0.657 \\
    3DGS+DG & 20.14 & 0.693 & 20.35 & 0.683 & 23.54 & 0.864 & 17.46 & 0.728 & 23.42 & 0.854 & 18.87 & 0.728 \\
    Ours & \textbf{21.15} & \textbf{0.737} & \textbf{21.01} & \textbf{0.701} & \textbf{26.42} & \textbf{0.897} & \textbf{21.63} & \textbf{0.827} & \textbf{26.21} & \textbf{0.907} & \textbf{22.87} & \textbf{0.837} \\
    \bottomrule
    \end{tabular}
}

%% file: supp/onthego2.tex
\resizebox{\textwidth}{!}{
    \begin{tabular}{l|cc|cc|cc|cc|cc|cc|cc}
    \toprule
    \multirow{2}{*}{Method}  
    & \multicolumn{2}{c|}{Arcdetriomphe}  
    & \multicolumn{2}{c|}{Drone} 
    & \multicolumn{2}{c|}{Statue}  
    & \multicolumn{2}{c|}{Train} 
    & \multicolumn{2}{c|}{Train-station} 
    & \multicolumn{2}{c|}{Tree}  
    & \multicolumn{2}{c}{Mean}  
    \\
     & PSNR & SSIM  
     & PSNR & SSIM  
     & PSNR & SSIM  
     & PSNR & SSIM  
     & PSNR & SSIM  
     & PSNR & SSIM  
     & PSNR & SSIM  
     \\
    \midrule
    3DGS~\cite{kerbl20233d}
    & 25.57 & 0.926 
    & \underline{21.37} & \underline{0.830} 
    & 15.95 & 0.751 
    & 22.49 & 0.847 
    & 21.43 & \textbf{0.871}
    & 22.44 & 0.846 
    & 21.54 & 0.845 
    \\
    SpotLessSplats~\cite{sabour2024spotlesssplats}
    & 28.70 & 0.940 
    & 20.87 & 0.800 
    & 16.01 & 0.737 
    & 23.28 & 0.841 
    & 21.37 & 0.815 
    & 23.00 & 0.834 
    & 22.21 & 0.828
    \\
    WildGaussians~\cite{kulhanek2024wildgaussians}
    & 24.25 & 0.898 
    & 21.31 & 0.815 
    & \textbf{17.32} & \underline{0.795} 
    & 23.81 & 0.852 
    & \underline{22.50} & 0.846 
    & 22.77 & 0.832 
    & 21.99 & 0.840 
    \\
    Robust3DGS~\cite{ungermann2024robust}
    & 26.36 & 0.933 
    & 18.69 & 0.785 
    & 14.66 & 0.724 
    & 23.79 & 0.860
    & 20.67 & 0.833 
    & 22.73 & \underline{0.868} 
    & 21.15 & 0.834 
    \\
    T-3DGS~\cite{pryadilshchikov2024t}
    & \underline{28.86} & \underline{0.943} 
    & 21.08 & 0.820 
    & 16.57 & 0.756 
    & \textbf{24.34} & \underline{0.870} 
    & 21.87 & \underline{0.851}
    & \underline{23.14} & \textbf{0.870} 
    & \underline{22.63} & \underline{0.852} 
    \\
    Ours 
    & \textbf{29.43} & \textbf{0.949} 
    & \textbf{21.62} & \textbf{0.844} 
    & \underline{16.65} & \textbf{0.802} 
    & \underline{24.07} & \textbf{0.871} 
    & \textbf{22.78} & \textbf{0.871} 
    & \textbf{23.57} & \underline{0.868} 
    & \textbf{23.02} & \textbf{0.868}
    \\
 
    \bottomrule
    \end{tabular}}

%% file: supp/vis_onthego2_1.tex
    \makebox[0.319\textwidth]{\small Robust3DGaussians}
    \makebox[0.319\textwidth]{\small SpotLessSplats}
    \makebox[0.319\textwidth]{\small WildGaussians}
    \\
    \includegraphics[width=0.319\textwidth]{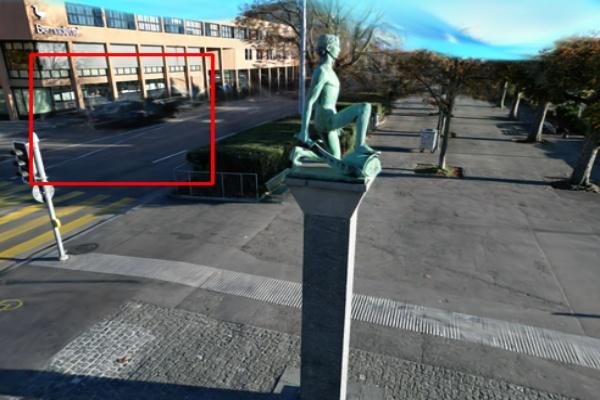}
    \includegraphics[width=0.319\textwidth]{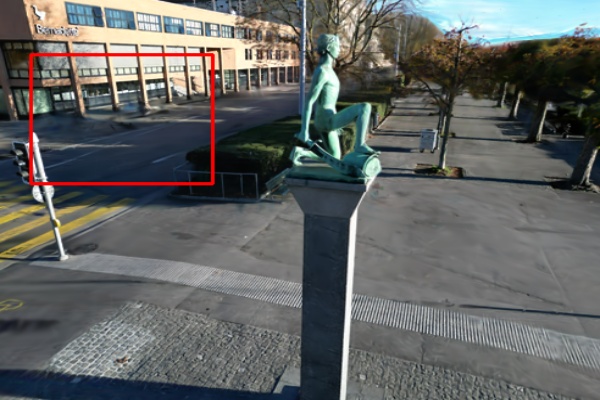}
    \includegraphics[width=0.319\textwidth]{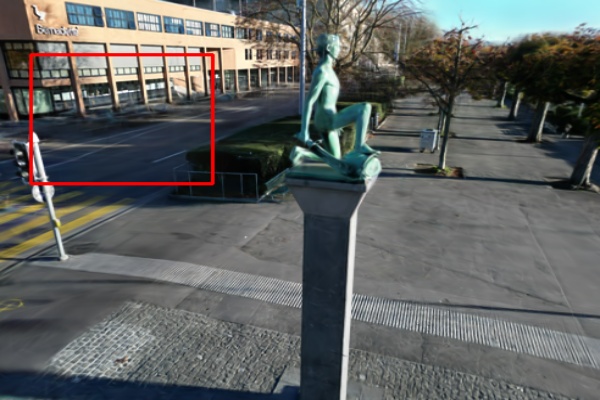}
    \\
    \includegraphics[width=0.319\textwidth]{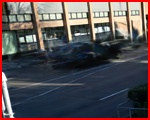}
    \includegraphics[width=0.319\textwidth]{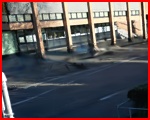}
    \includegraphics[width=0.319\textwidth]{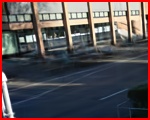}
    \\
    \makebox[0.319\textwidth]{\small T-3DGS}
    \makebox[0.319\textwidth]{\small Ours}
    \makebox[0.319\textwidth]{\small Ground Truth}
    \\
    \includegraphics[width=0.319\textwidth]{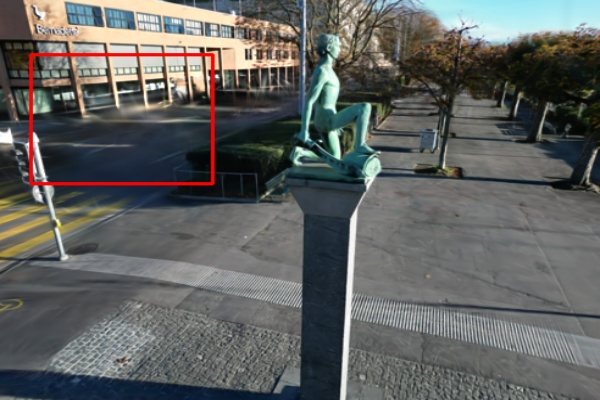}
    \includegraphics[width=0.319\textwidth]{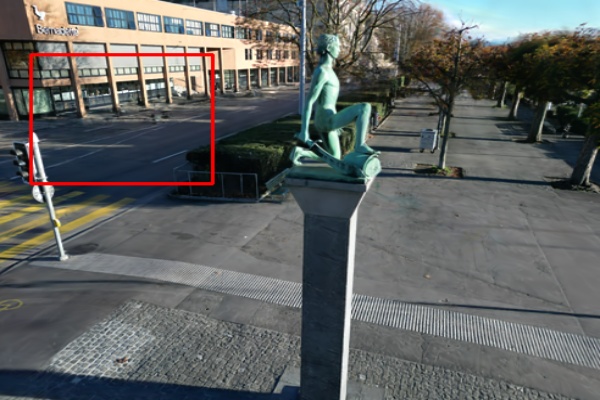}
    \includegraphics[width=0.319\textwidth]{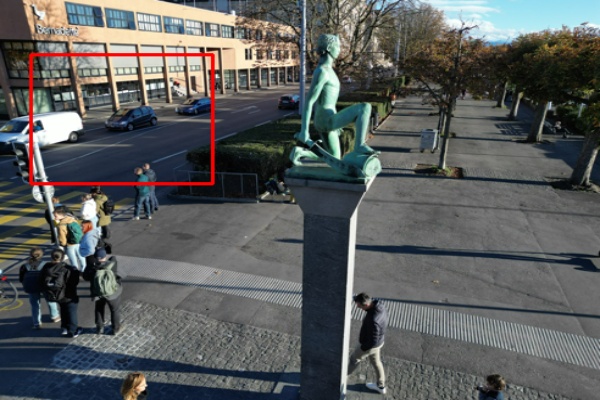}
    \\
    \includegraphics[width=0.319\textwidth]{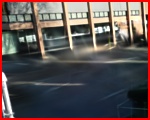}
    \includegraphics[width=0.319\textwidth]{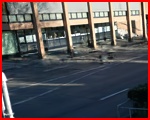}
    \includegraphics[width=0.319\textwidth]{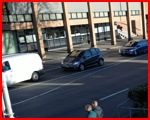}
    \\

%% file: supp/vis_onthego2_2.tex
    \makebox[0.319\textwidth]{\small Robust3DGaussians}
    \makebox[0.319\textwidth]{\small SpotLessSplats}
    \makebox[0.319\textwidth]{\small WildGaussians}
    \\
    \includegraphics[width=0.319\textwidth]{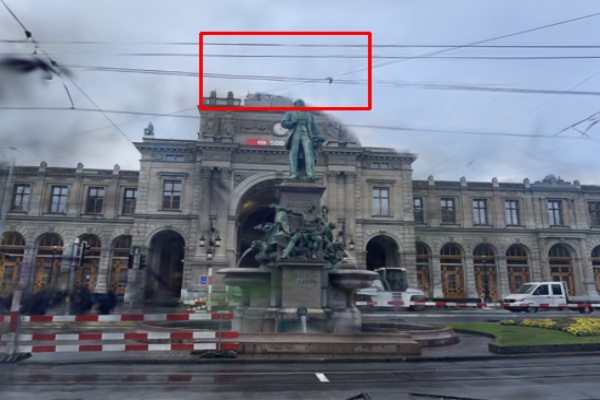}
    \includegraphics[width=0.319\textwidth]{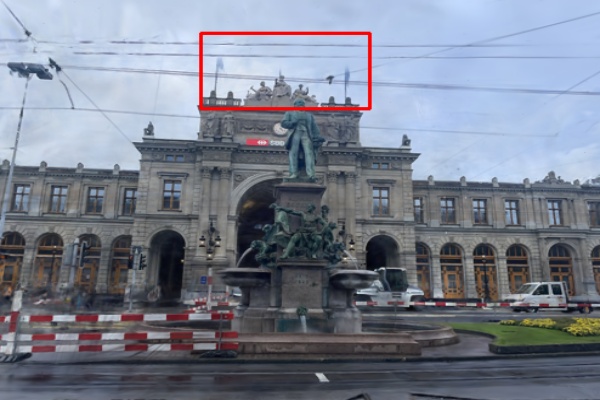}
    \includegraphics[width=0.319\textwidth]{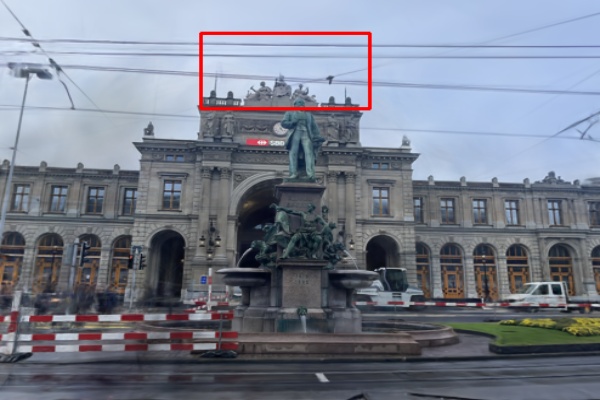}
    \\
    \includegraphics[width=0.319\textwidth]{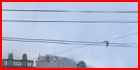}
    \includegraphics[width=0.319\textwidth]{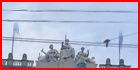}
    \includegraphics[width=0.319\textwidth]{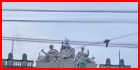}
    \\
    \makebox[0.319\textwidth]{\small T-3DGS}
    \makebox[0.319\textwidth]{\small Ours}
    \makebox[0.319\textwidth]{\small Ground Truth}
    \\
    \includegraphics[width=0.319\textwidth]{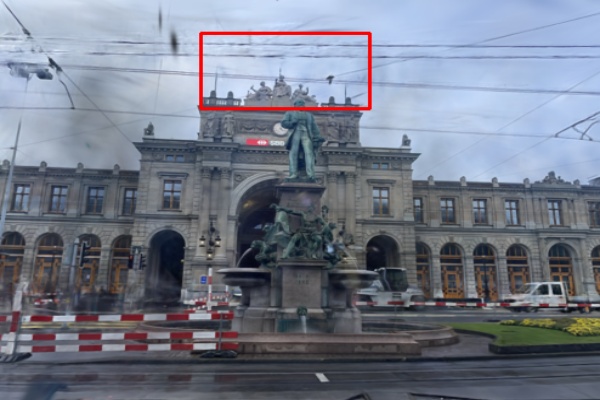}
    \includegraphics[width=0.319\textwidth]{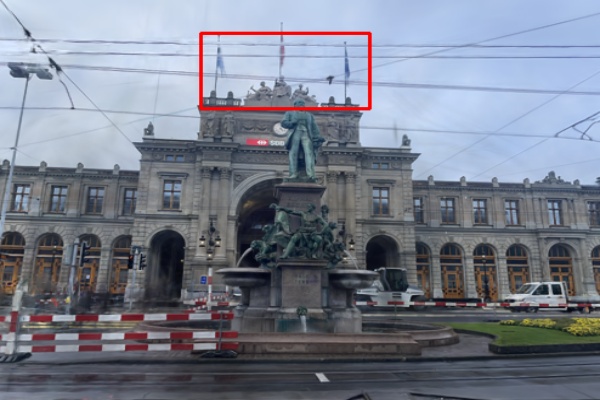}
    \includegraphics[width=0.319\textwidth]{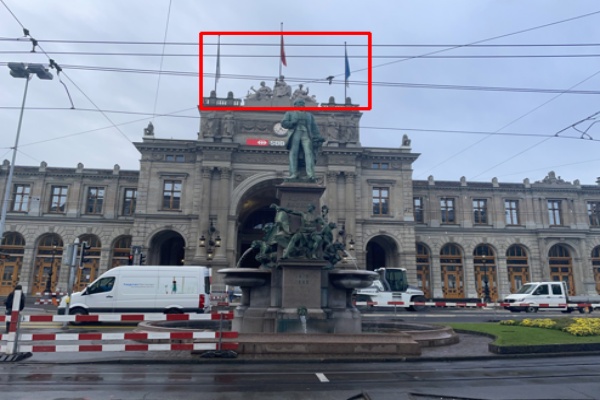}
    \\
    \includegraphics[width=0.319\textwidth]{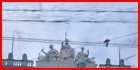}
    \includegraphics[width=0.319\textwidth]{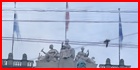}
    \includegraphics[width=0.319\textwidth]{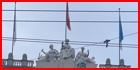}
    \\

%% file: supp/vis_onthego2_3.tex
    \makebox[0.319\textwidth]{\small Robust3DGaussians}
    \makebox[0.319\textwidth]{\small SpotLessSplats}
    \makebox[0.319\textwidth]{\small WildGaussians}
    \\
    \includegraphics[width=0.319\textwidth]{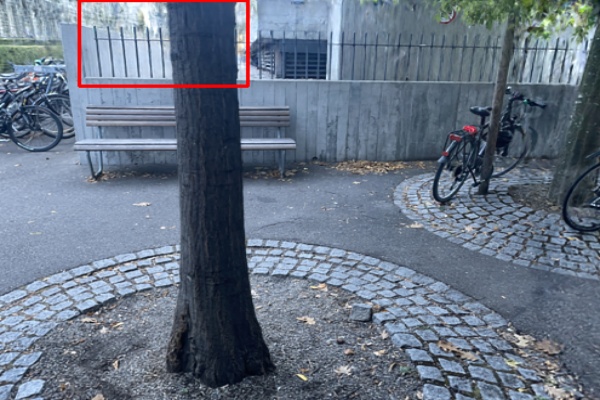}
    \includegraphics[width=0.319\textwidth]{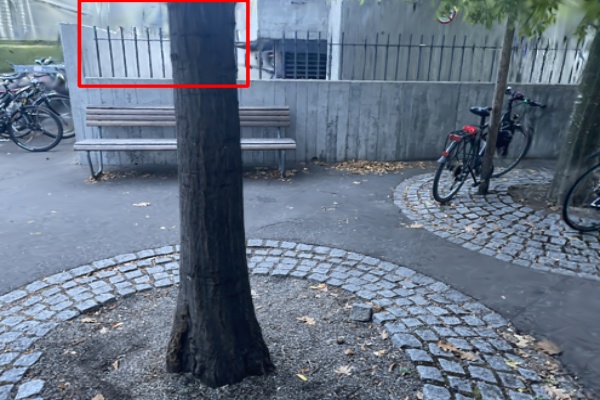}
    \includegraphics[width=0.319\textwidth]{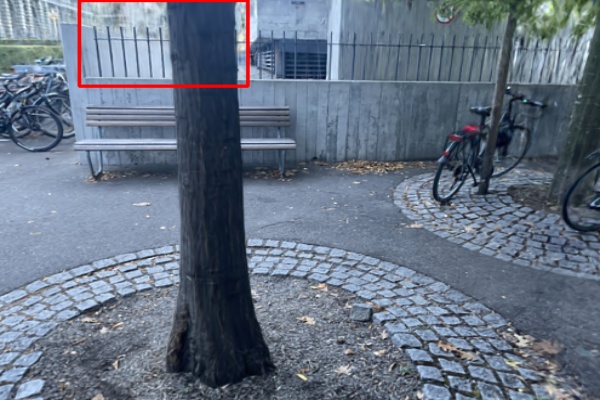}
    \\
    \includegraphics[width=0.319\textwidth]{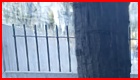}
    \includegraphics[width=0.319\textwidth]{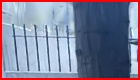}
    \includegraphics[width=0.319\textwidth]{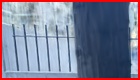}
    \\
    \makebox[0.319\textwidth]{\small T-3DGS}
    \makebox[0.319\textwidth]{\small Ours}
    \makebox[0.319\textwidth]{\small Ground Truth}
    \\
    \includegraphics[width=0.319\textwidth]{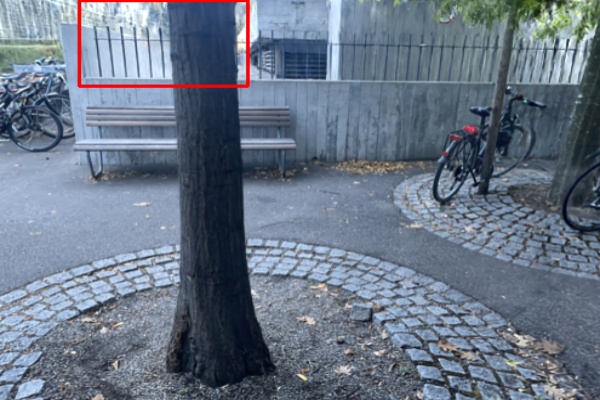}
    \includegraphics[width=0.319\textwidth]{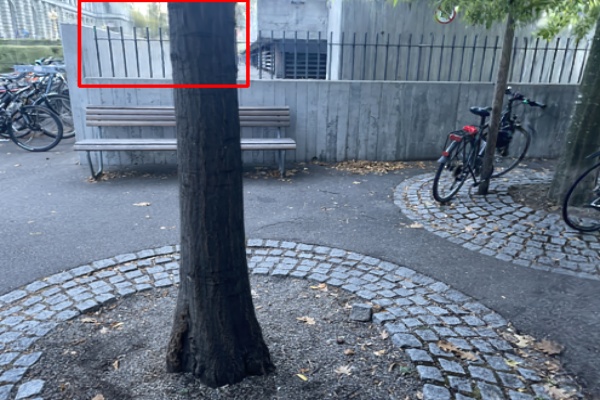}
    \includegraphics[width=0.319\textwidth]{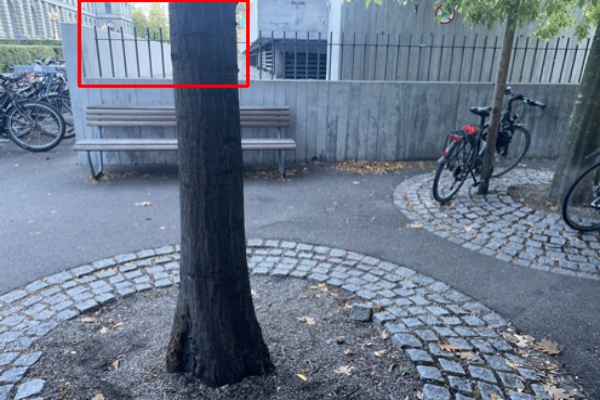}
    \\
    \includegraphics[width=0.319\textwidth]{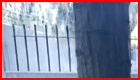}
    \includegraphics[width=0.319\textwidth]{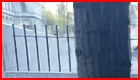}
    \includegraphics[width=0.319\textwidth]{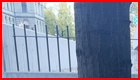}
    \\